\documentclass[10pt,twocolumn,letterpaper]{article}

\usepackage[pagenumbers]{cvpr} 

\usepackage{graphicx}
\usepackage{amsmath}
\usepackage{amssymb}
\usepackage{booktabs}
\usepackage{rotating}
\usepackage{array}
\usepackage{colortbl}
\usepackage{multirow}
\usepackage{color}
\usepackage{enumitem}
\usepackage[dvipsnames]{xcolor}
\usepackage[ruled,linesnumbered]{algorithm2e}
\DeclareMathOperator*{\diag}{diag}
\usepackage{multicol}
\usepackage[pagebackref,breaklinks,colorlinks,citecolor=RoyalBlue]{hyperref}

\usepackage[capitalize]{cleveref}
\crefname{section}{Sec.}{Secs.}
\Crefname{section}{Section}{Sections}
\Crefname{table}{Table}{Tables}
\crefname{table}{Tab.}{Tabs.}

\def\cvprPaperID{1590} 
\def\confName{CVPR}
\def\confYear{2022}

\begin{document}
	
	\title{Class-Balanced Pixel-Level Self-Labeling for \\ Domain Adaptive Semantic Segmentation}
	
	\author{Ruihuang Li$^1$,\quad Shuai Li$^1$,\quad Chenhang He$^1$,\quad Yabin Zhang$^1$,\quad Xu Jia$^2$,\quad Lei Zhang$^1$\thanks{Corresponding Author}\\
	$^1$The Hong Kong Polytechnic
University,\quad $^2$Dalian University of Technology\\
{\tt\small \{csrhli, csshuaili, csche, csybzhang, cslzhang\}@comp.polyu.edu.hk}, {\tt\small xjia@dlut.edu.cn} 
	}
	
	
	\maketitle
	
	\begin{abstract}
		Domain adaptive semantic segmentation aims to learn a model with the supervision of source domain data, and produce satisfactory dense predictions on unlabeled target domain. One popular solution to this challenging task is self-training, which selects high-scoring predictions on target samples as pseudo labels for training. However, the produced pseudo labels often contain much noise because the model is biased to source domain as well as majority categories. To address the above issues, we propose to directly explore the intrinsic pixel distributions of target domain data, instead of heavily relying on the source domain. Specifically, we simultaneously cluster pixels and rectify pseudo labels with the obtained cluster assignments. This process is done in an online fashion so that pseudo labels could co-evolve with the segmentation model without extra training rounds. To overcome the class imbalance problem on long-tailed categories, we employ a distribution alignment technique to enforce the marginal class distribution of cluster assignments to be close to that of pseudo labels. The proposed method, namely Class-balanced Pixel-level Self-Labeling (CPSL), improves the segmentation performance on target domain over state-of-the-arts by a large margin, especially on long-tailed categories. The source code is available at \url{https://github.com/lslrh/CPSL}. 
		
	\end{abstract}
	\section{Introduction}
	Semantic segmentation is a fundamental computer vision task, which aims to make dense semantic-level predictions on images~\cite{long2015fully,zhao2017pyramid,chen2017deeplab,lin2017refinenet,wang2021exploring}. It is a key step in numerous applications, including autonomous driving, human-machine interaction, and augmented reality, to name a few. In the past few years, the rapid development of deep Convolutional Neural Networks (CNNs) has boosted semantic segmentation significantly in terms of accuracy and efficiency. However, the performance of deep models trained in one domain often drops largely when they are applied to unseen domains. For example, in autonomous driving the segmentation model is confronted with great challenges when weather conditions are changing constantly~\cite{zou2018unsupervised}. A natural way to improve the generalization ability of segmentation model is to collect data from as many scenarios as possible. However, it is very costly to annotate pixel-wise labels for a large amount of images ~\cite{cordts2016cityscapes}. More effective and practical approaches are required to address the domain shifts of semantic segmentation. 
	
	Unsupervised Domain Adaptation (UDA) provides an important way to transfer the knowledge learned from one labeled source domain to another unlabeled target domain. For example, we can collect many synthetic data whose dense annotations are easy to get by using game engines such as GTA5~\cite{richter2016playing} and SYNTHIA~\cite{ros2016synthia}. Then the question turns to how to adapt the model trained from a labeled synthetic domain to an unlabeled real image domain. Most previous works of UDA bridge the domain gap by aligning data distributions at the image level~\cite{hoffman2017cycada,li2019bidirectional,murez2018image}, feature level~\cite{chen2019progressive,hoffman2017cycada,hoffman2016fcns,li2021t} or output level~\cite{melas2021pixmatch,luo2019taking,tsai2018learning}, through adversarial training or auxiliary style transfer networks. However, these techniques will increase the model complexity and make the training process unstable, which impedes their reproducibility and robustness. 
	
	Another important approach is self-training~\cite{zou2018unsupervised,zou2019confidence,zhang2017curriculum}, which alternatively generates pseudo labels by selecting high-scoring predictions on target domain and provides supervision for the next round of training. Though these methods have produced promising performance, there are still some major limitations. On one hand, the segmentation model tends to be biased to source domain so that the pseudo labels produced on target domain are error-prone; on the other hand, highly-confident predictions may only provide very limited supervision information for the model training. To solve these issues, some methods~\cite{zhang2019category,zhang2021prototypical} have been proposed to produce more accurate and informative pseudo labels. For example, instead of using the classifier trained on source domain to generate pseudo labels, Zhang \etal~\cite{zhang2019category} assigned pseudo labels to pixels based on their distances to the category prototypes. These prototypes, however, were built in source domain and usually deviated much from the target domain. ProDA~\cite{zhang2021prototypical} leveraged the feature distances from prototypes to perform online rectification, but it was challenging to construct prototypes for long-tailed categories, which often led to unsatisfactory performance.  
	
	Different from previous self-training methods which use classifier-based noisy pseudo labels for supervision, in this paper we propose to perform online pixel-level self-labeling via clustering on target domain, and use the resulting soft cluster assignments to correct pseudo labels. Our idea comes from the fact that pixel-wise cluster assignments could reveal the intrinsic distributions of pixels in target domain, and provide useful supervision for model training. Compared to conventional label generation methods that are often biased towards source domain, cluster assignment in target domain is more reliable as it explores inherent data distribution. Considering that the classes of segmentation dataset are highly imbalanced (please refer to Fig.~\ref{fig2}), we employ a distribution alignment technique to enforce the class distribution of cluster assignments to be close to that of pseudo labels, which is more favorable to class-imbalanced dense prediction tasks. The proposed Class-balanced Pixel-level Self-Labeling (CPSL) module works in a plug-and-play fashion, which could be seamlessly incorporated into existing self-training framework for UDA. The major contributions of this work are summarized as follows:\\
	\vspace{-0.5em}
	\vspace{-1.0em}\begin{itemize}[leftmargin=*]
		\item[$\bullet$] A pixel-level self-labeling module is developed for domain adaptive semantic segmentation. We cluster pixels in an online fashion and simultaneously rectify pseudo labels based on the resulting cluster assignments.\vspace{-0.5em}
		\item[$\bullet$] A distribution alignment technique is introduced to align the class distribution of cluster assignments to that of pseudo labels, aiming to improve the performance over long-tailed categories. A class-balanced sampling strategy is adopted to avoid the dominance of majority categories in pseudo label generation. \vspace{-0.5em}
		\item[$\bullet$] Extensive experiments demonstrate that the proposed CPSL module improves the segmentation performance on target domain over state-of-the-arts by a large margin. It especially shows outstanding results on long-tailed classes such as ``motorbike'', ``train'', ``light'', \etc.
	\end{itemize}\vspace{-1.0em}

	\begin{figure*}
		\centering
		\includegraphics[scale=0.19]{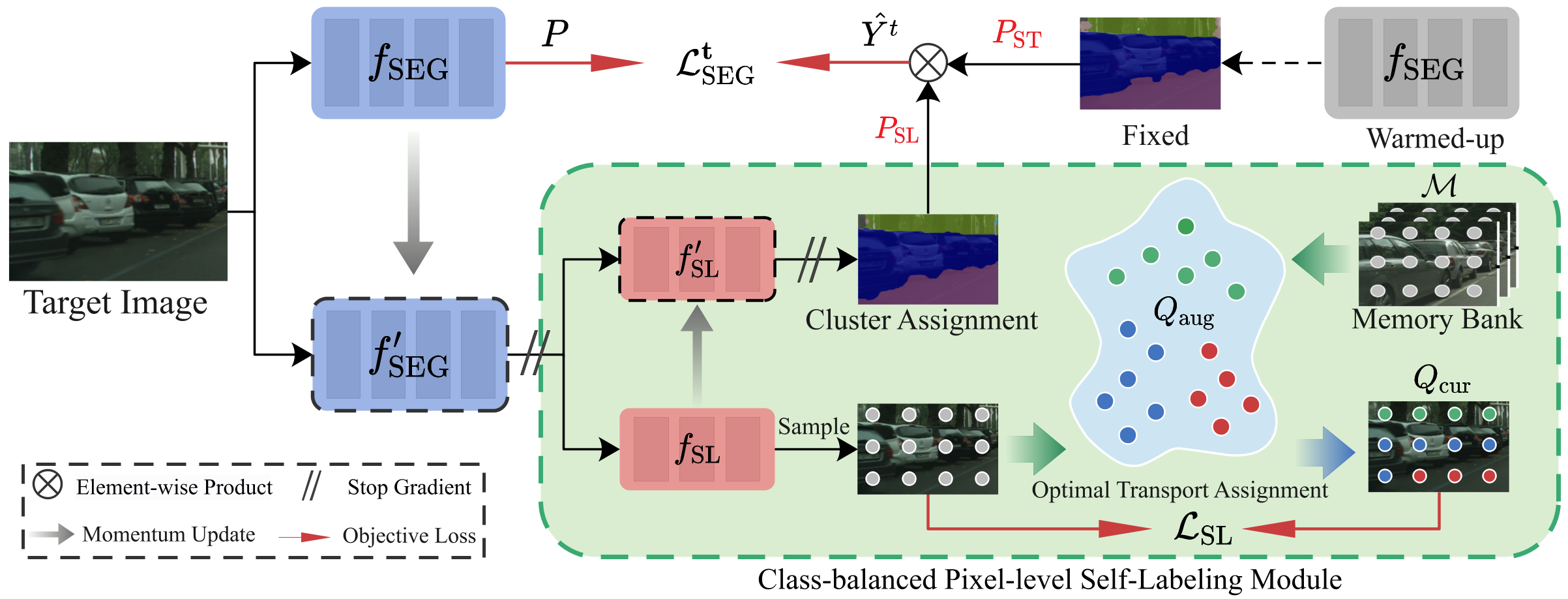}\\
		\caption{The framework of Class-balanced Pixel-level Self-Labeling (CPSL). The model contains a main segmentation network $f_{\rm SEG}$ and its momentum-updated version $f'_{\rm SEG}$. The $f'_{\rm SEG}$ is followed by a self-labeling head $f_{\rm{SL}}$ and its momentum version $f_{\rm{SL}}'$, which projects pixel-wise feature embedding into a class probability vector. The pixel-level self-labeling module produces soft cluster assignment $P_{\rm SL}$ to gradually rectify soft pseudo label $P_{\rm ST}$. Then the segmentation loss $\mathcal{L}_{\rm SEG}^t$ is computed between the prediction map $P$ and the rectified pseudo label $\hat{Y}^t$. To train the self-labeling head, we randomly sample pixels from each image, and use the memory bank $\mathcal{M}$, which contains previous batches of pixel features, to augment the current batch. Then we compute the optimal transport assignment $Q_{aug}$ over the augmented data by enforcing class balance, and use the assignment of current batch $Q_{cur}$ to compute the self-labeling loss $\mathcal{L}_{\rm SL}$.}
		\label{0fig1}
		\vspace{-0.5em}
	\end{figure*}
	\section{Related Work}
	\noindent{\bf Semantic Segmentation.} The goal of semantic segmentation is to segment an image into regions of different semantic categories. While the Fully Convolutional Networks (FCNs) \cite{long2015fully} have greatly boosted the performance of semantic segmentation, they have relatively small receptive field to explore visual context. Many later works focus on how to enlarge the receptive field of FCNs to model long-range context dependencies of images, such as dilated convolution~\cite{chen2017deeplab}, multi-layer feature fusion~\cite{lin2017refinenet}, spatial pyramid pooling~\cite{zhao2017pyramid} and variants of non-local blocks~\cite{fu2019dual,huang2019ccnet,hu2018squeeze}.
	However, directly applying these models to unseen domains will induce poor segmentation performance because of their weak generalization ability. Therefore, many domain adaptation techniques have been proposed to improve model generalization ability on new domains. 
	
	\noindent{\bf Domain Adaptation for Semantic Segmentation.} Recently, many works have been proposed to bridge the domain gap and improve the adaptation performance. The most representative ones are adversarial training-based methods \cite{kim2020learning,hong2018conditional,pan2020unsupervised,tsai2018learning,tsai2019domain}, which aim to align different domains on intermediate features or network predictions. Style transfer-based methods~\cite{chang2019all,chen2019crdoco,choi2019self,wu2018dcan,yang2020fda} minimize domain gap at the image level. For example, Chang~\etal~\cite{chang2019all} proposed to disentangle an image into domain-invariant structures and domain-specific textures for image translation. The training process of these models is rather complex since multiple networks, such as discriminators or style transfer networks, have to be trained concurrently.\\
	\indent Another important technique for UDA is self-training~\cite{zou2018domain,zou2019confidence,lian2019constructing,zhang2019category,melas2021pixmatch,li2021t}, which iteratively generates pseudo labels on target data for model update. Zou~\etal~\cite{zou2018domain} proposed a class-balanced self-training method for domain adaption of semantic segmentation. To reduce the noise in pseudo labels, Zou~\etal~\cite{zou2019confidence} further proposed a confidence regularized self-training method, which treated pseudo labels as trainable latent variables. Lian~\etal~\cite{lian2019constructing} constructed a pyramid curriculum for exploring various properties about the target domain. Zhang~\etal~\cite{zhang2019category} enforced category-aware feature alignment by choosing the prototypes of source domain as guided anchors. ProDA~\cite{zhang2021prototypical} went further by employing the feature distances from each pixel to prototypes to correct pseudo labels pre-computed by the source model. These methods, however, neglect either the pixel-wise intrinsic structures or inherent class distribution of target domain images, tending to be biased to source domain or majority classes.
	
	\noindent {\bf Clustering-based Representation Learning.} Our work is also related to clustering-based methods~\cite{asano2020self,bautista2016cliquecnn,caron2018deep,huang2019unsupervised,xie2016unsupervised,yan2020clusterfit,yang2016joint,zhuang2019local,asano2020labelling}. Caron~\etal~\cite{caron2018deep} iteratively performed $k$-means on latent representations and used the produced cluster assignments to update network parameters. Recently, Asano~\etal~\cite{asano2020self} cast the cluster assignment problem as an optimal transport problem which can be solved efficiently through a fast variant of the Sinkhorn-Knopp algorithm. SwAV~\cite{caron2020unsupervised} performed clustering while enforcing consistency among the cluster assignments of different augmentations of the same image. In this paper, we extend self-labeling from image-level classification to pixel-level semantic segmentation. In addition, different from Asano~\etal~\cite{asano2020self} and Caron~\etal~\cite{caron2018deep}, we compute cluster assignments in an online fashion, making our method scalable to dense pixel-wise prediction tasks. 
	
	\section{Method}
	\subsection{Overall Framework}
	\label{sec3.1}
	In the setting of unsupervised domain adaptation for semantic segmentation, we are provided with a set of labeled data in source domain $\mathcal{D}_{S}=\{(X^s_n,Y^s_n)\}^{N_S}_{n=1}$, where $X^s_n$ is the source image with label $Y^s_n$ and $N_S$ is the number of images, as well as a set of $N_T$ unlabeled images $X^t_n$ in target domain $\mathcal{D}_{T}=\{X^t_n\}^{N_T}_{n=1}$. Both domains share the same $C$ classes. Our goal is to learn a model by using the labeled source data in $\mathcal{D}_{S}$ and unlabeled target data in $\mathcal{D}_{T}$, which could perform well on unseen test data in the target domain.
	
	The overall framework of our proposed CPSL is shown in Fig.~\ref{0fig1}. We propose a pixel-level self-labeling module (highlighted in the green color box) to explore the intrinsic pixel-wise distributions of the target domain data via clustering, and to reduce the noise in pseudo labels. Before the training, we first generate a soft pseudo label map $P_{\rm ST}\in \mathbb{R}^{H\times W\times C}$ for each target domain image by a warmed-up model that is pre-trained on the source domain data. The obtained $P_{\rm ST}$ is usually error-prone because of the large domain shift. Therefore, in the training process, we rectify $P_{\rm ST}$ incrementally with the soft cluster assignment, denoted by $P_{\rm SL}\in \mathbb{R}^{H\times W\times C}$. Specifically, the rectification of $P_{\rm ST}$ is conducted as follows:
	{\small 	\begin{align}
			\label{eq1}
			\hat{Y}^{t,(c)}_{n,i} =\begin{cases}
				1, & \text{ if } c=\underset{c*}{argmax}(P^{(c*)} _{{\rm SL},n,i}\cdot P^{(c*)} _{{\rm ST},n,i}) \\ 
				0, & \text{ otherwise }  
			\end{cases},
	\end{align}}where $\hat{Y}_{n,i}^{t,(c)}$ denotes the $c$-th element of rectified pseudo label at the $i$-th pixel of target image $X^t_n$. $P^{(c*)} _{{\rm SL},n,i}$ represents the probability that the $i$-th pixel of $X^t_n$ belongs to the $c*$-th category. Eq.~\ref{eq1} has a similar formulation to~\cite{qi2018low,snell2017prototypical,zhang2021prototypical}, where $P_{{\rm SL}}$ can be regarded as the weight map to modulate the softmax probability map $P_{\rm ST}$. The cluster assignment $P_{\rm SL}$ exploits the inherent data distribution of target domain, thus it is highly complementary to the classifier-based pseudo label $P_{\rm ST}$ which heavily relies on source domain.
	
	We define the segmentation loss on target domain, denoted by $\mathcal{L}_{\rm SEG}^t$, as the pixel-level cross-entropy loss between the segmentation probability map $P_n\in \mathbb{R}^{H\times W\times C}$ and the rectified pseudo label $\hat{Y}_n^t$ of target image $X^t_{n}$:
	\begin{align}
		\mathcal{L}_{\rm SEG}^t = -\sum_{n=1}^{N_T}\sum_{i=1}^{H\times W}\sum_{c=1}^{C}\hat{Y}^{t,(c)}_{n,i} \log{P}_{n,i}^{(c)}.
		\label{eq2}
	\end{align}   
	In addition, the loss on source domain, denoted by $\mathcal{L}_{\rm SEG}^s$, can be defined as the standard pixel-wise cross-entropy on the labeled images:
	\begin{align}
		\mathcal{L}_{\rm SEG}^s = -\sum_{n=1}^{N_S}\sum_{i=1}^{H\times W}\sum_{c=1}^C Y_{n,i}^{s,(c)} \log P_{n,i}^{(c)}.
		\label{eq3}
	\end{align}
	Then the total segmentation loss $\mathcal{L}_{\rm SEG}$ is obtained as the sum of them: $\mathcal{L}_{\rm SEG} = \mathcal{L}^t_{\rm SEG}+\mathcal{L}^s_{\rm SEG}$.
	
	In the following subsections, we will explain in detail the design of our CPSL module.
	\begin{figure}
		\centering
		\includegraphics[scale=0.15]{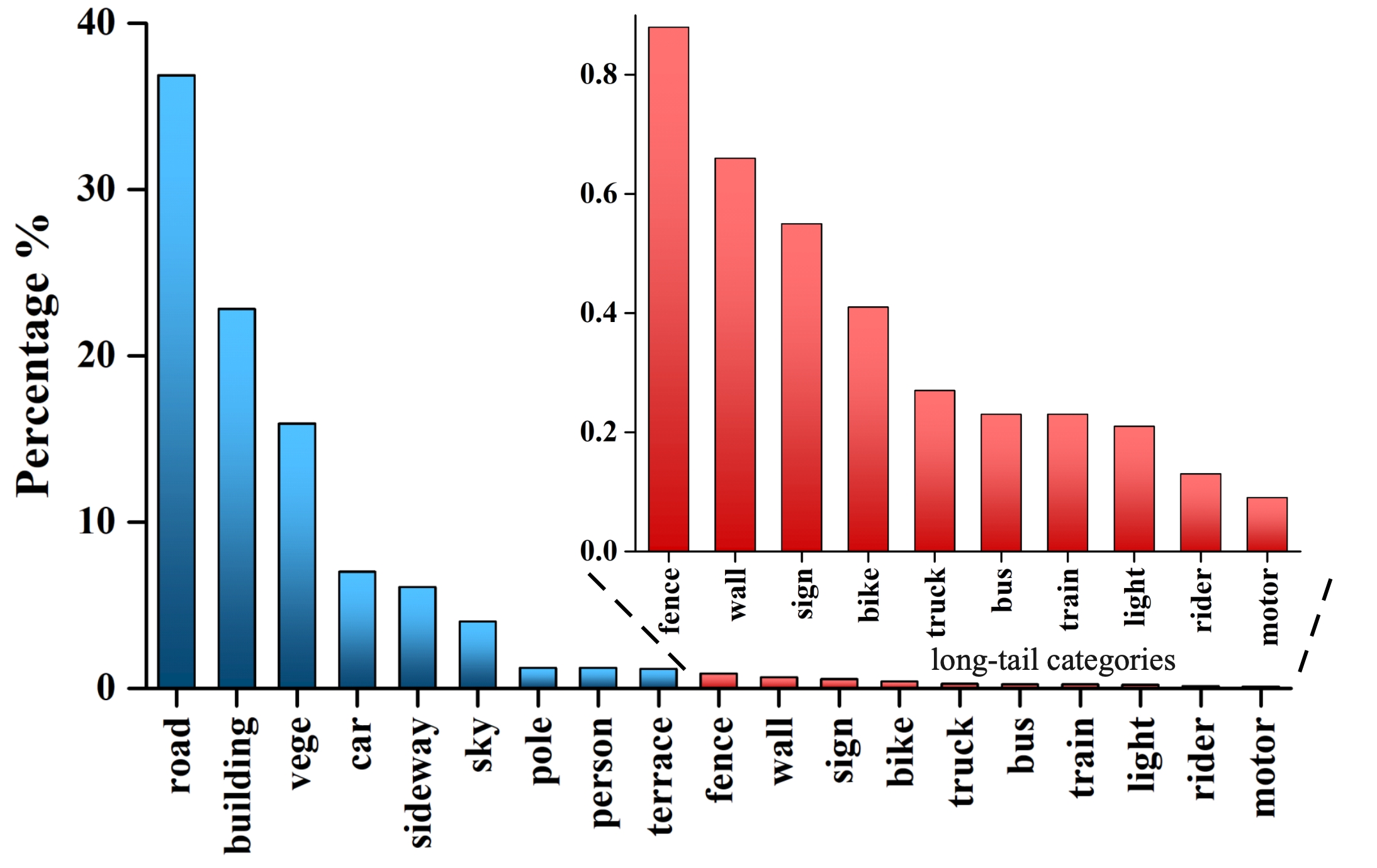}\\\vspace{-1.0em}
		\caption{The class distribution of the Cityscapes dataset. }
		\label{fig2}
		\vspace{-1.0em}
	\end{figure}
	\subsection{Online Pixel-Level Self-Labeling}
	\label{sec3.2}
	\noindent{\bf Pixel-Level Self-Labeling.} Conventional self-training based methods usually use a model pre-trained on source domain to produce pseudo labels, which often contain much noise~\cite{zou2018domain,zou2019confidence,zhang2019category}. To clean the pseudo labels, we propose to perform pixel-level self-labeling via clustering on target domain and use the obtained cluster assignments to rectify the pseudo labels. The basic motivation is that pixel-wise clustering could reveal the intrinsic structures of target domain data, and it is complementary to the classifier trained on source domain data. Thus, cluster assignments could provide extra supervision for training a domain adaptive segmentation model. 
	
	Specifically, we first extract features from an input image to obtain ${Z}\in \mathbb{R}^{H\times W\times D}$ and normalize it with ${{ z}_i}=\frac{{ z}_i}{||{ z}_i||_2}$, where $z_i$ is the $i$-th feature vector of $Z$ with length $D$. Then we randomly sample a group of pixels $\hat{Z}=[{ z}_1,\cdots,{ z}_M]$ from each image, and pass them through a self-labeling head $f_{\rm SL}$. Finally, we obtain their class probability vectors $\hat{P}= [p_1,\cdots,p_M]$ by taking a softmax operation: 
	\begin{align}
		p_m^{(c)} = \frac{\exp(\frac{1}{\tau}f_{\rm SL}^{(c)}(z_m))}{{\sum_{c'}\exp(\frac{1}{\tau}f_{\rm SL}^{(c')}(z_m))}}, \; c \in \{1,\cdots,C\},
	\end{align}
	where $f_{\rm SL}^{(c)}(z_m)$ is the $c$-th element of the output of $z_m$ from self-labeling head. $p_m^{(c)}$ denotes the probability that the $m$-th pixel belongs to the $c$-th category. $\tau$ is a temperature parameter. Considering there is no ground truth label available for target data, we train the head $f_{\rm SL}$ through a self-labeling mechanism \cite{asano2020self} with the following objective function:
	{\small 
		\begin{equation}
			\begin{split}
				&\mathcal{L}_{\rm SL}=-\frac{1}{M}\sum_{m=1}^{M}\sum_{c=1}^{C}{ q}_m^{(c)}\log{ p}_m^{(c)}\quad  s.t.\, Q\in\mathbb{Q},
				\\ & {\rm with}\quad  \mathbb{Q}:=\{Q\in \mathbb{R}_+^{C\times M}|Q\mathbf{1}_M=r, Q^T\mathbf{1}_C=h\}.
				\label{eq4}
	\end{split}\end{equation}}The above formula is an instance of the optimal transport problem~\cite{cuturi2013sinkhorn}, where $Q=\frac{1}{M}[{ q}_1, \cdots, { q}_M]$ is a transport assignment and it is restricted to be a probability matrix by satisfying the constraint $\mathbb{Q}$.  $\mathbf{1}_C$ and $\mathbf{1}_M$ denote the vectors of ones with dimension $C$ and $M$, respectively. $r$ and $h$ are the marginal projections of $Q$ onto its rows and columns, respectively. 
	
	By formulating the cluster assignment problem as an optimal transport problem, the optimization of Eq.~\ref{eq4} with respect to variable $Q$ can be solved efficiently by the iterative Sinkhorn-Knopp algorithm~\cite{cuturi2013sinkhorn}. The optimal solution is obtained by:
	\begin{align}
		Q^* = {\diag}(\alpha)\exp(\frac{f_{\rm SL}(\hat{Z})}{\varepsilon})\diag(\beta),
	\end{align}
	where $\alpha \in \mathbb{R}^C$ and $\beta \in \mathbb{R}^M$ are two renormalization vectors which can be computed efficiently in linear time even for dense prediction tasks. $\varepsilon$ is a temperature parameter. 
	
	Then by fixing label assignment $Q$, the self-labeling head $f_{\rm SL}$ is updated by minimizing $\mathcal{L}_{\rm SL}$ with respect to $\hat{P}$, which is the same as training with cross-entropy loss.
	
	\vspace{0.5em}\noindent\textbf{Weight Initialization.} We use the soft cluster assignment $P_{\rm SL}$ to rectify the classifier-based pseudo label $P_{\rm ST}$. However, the clustering categories usually mismatch those of the classifier, resulting in performance degradation. To overcome this issue, we initialize the weight of self-labeling head $f_{\rm SL}$ with category prototypes. Specifically, we compute the prototypes $[\bar{\bf{z}}_1,\cdots,\bar{\bf{z}}_C]$ for each category through:
	\begin{equation}
		\bar{\mathbf{z}}_c=\frac{1}{|\Gamma_c|}\sum_{n=1}^{N_T}\sum_{i=1}^{H\times W}Y_{{\rm ST},n,i}^{(c)}\cdot z_{n,i},
	\end{equation}
	where $|\Gamma_c|$ denotes the number of pixels belonging to the $c$-th category in all images. $Y_{{\rm ST}}$ is the hard version of $P_{{\rm ST}}$. Then the self-labeling process can be regarded as assigning pixels to different prototypes. In this way, the clustering categories are able to match classification categories.
	
	\vspace{0.5em}\noindent {\bf{Online Cluster Assignment.}} Different from Asano~\etal~\cite{asano2020self}, where the assignment $Q$ is computed over the full dataset, we conduct online clustering on data batches during training. Considering that the number of samples in a mini-batch is often too small to cover all categories, and the class distribution varies largely across different batches, we augment the features $\hat{Z}$ with a memory bank $\mathcal{M}$, which is updated on-the-fly, to reduce the randomness of sampling. Specifically, throughout the training process, we maintain a queue of 65,536 pixel features from previous batches in $\mathcal{M}$. In each iteration, we compute the optimal transport assignment on the augmented data $Z_{aug}$, denoted by $Q_{aug}$, but only the assignment of current batch, denoted by $Q_{cur}$, is used to compute the self-labeling loss $\mathcal{L}_{\rm SL}$. In this way, we could alternatively update the self-labeling head $f_{\rm SL}$ and use it to generate more accurate cluster assignment $P_{\rm SL}$ online. Hence, the pseudo labels will be improved incrementally by the resulting cluster assignments, and the noise will be gradually reduced without extra rounds of training. 
	
	\subsection{Class-Balanced Self-Labeling}
	\label{sec3.3}
	As shown in Fig.~\ref{fig2}, there exists severe class-imbalance in current semantic segmentation datasets. Some long-tailed classes have very limited pixels (\eg, ``traffic light", ``sign"), and some classes only appear in a few images (\eg, ``motorbike'', ``train"). Such a problem will make it difficult to train a robust segmentation model, especially for those long-tailed classes. In this work, we propose two techniques to address this issue, \ie, class-balanced sampling and distribution alignment.
	
	\vspace{0.5em}\noindent {\bf Class-Balanced Sampling.} 
	We randomly sample pixels from each image, which makes the class distribution of data in memory bank $\mathcal{M}$ approach to that of the whole dataset. In order to make sure that the pixels of long-tailed categories can be selected equally, we sample from different categories with the same proportion, \ie, $\frac{M}{H\times W}$, where $M$ is the number of pixels to be sampled in each image. For each input image $X_n^t$, we first compute its class distribution $\delta_n$ through 
	\begin{align}
		\label{eq7}
		\delta_n^{(c)}=\frac{1}{ H\times W}\sum_{i}^{H\times W}\hat{Y}^{t,(c)}_{n,i},
	\end{align}
	where $\delta^{(c)}_n$ denotes the proportion of pixels belonging to the $c$-th category in image $X_n^t$. Then the number of samples $M_c$ for each category $c$ is decided by: 
	\begin{align}
		M_c = \left \lfloor M\times \delta_n^{(c)}\right \rfloor.
	\end{align} 
	If image $X_n^t$ does not contain certain classes of pixels, we will randomly sample the rest pixels from other categories to make up $M$ samples.
	
	\vspace{0.5em}\noindent {\bf{Distribution Alignment.}} 
	As discussed in~\cite{asano2020self,caron2018deep}, simultaneously optimizing $Q$ and $\hat{P}$ in Eq.~\ref{eq4} may lead to degenerated results that all data points are trivially assigned to a single cluster. To avoid this, Asano~\etal~\cite{asano2020self} constrained that $Q$ should induce an equipartition of the data. However, this constraint is not reasonable and it will degrade the performance if the ground truth class distribution of the data, denoted by $\delta_{gt}$, is not uniform. In the Cityscapes dataset~\cite{cordts2016cityscapes}, for example, the number of pixels of the largest category (``road'') is approximately 300 times that of the smallest category (``motorbike''). 
	
	To overcome this problem, we propose a novel technique, namely distribution alignment, to align the distribution of cluster assignments to ground truth class distribution $\delta_{gt}$, aiming at partitioning pixels into subsets of unequal sizes. However, $\delta_{gt}$ is unknown since the true labels of target domain data are unavailable. Thus we propose to employ the moving average of pseudo labels' class distribution ${\delta}_{pseudo}$ to approximate ${\delta}_{gt}$. Specifically, we first initialize $\delta_{pseudo}$ based on the fixed pseudo labels $Y^t_{\rm ST}$ as follows:
	\begin{align}
		\delta_{pseudo}^{(c)}|_0=\frac{1}{N_T\times H\times W}\sum_{n}^{N_T}\sum_{i}^{H\times W}{ Y^{t,(c)}_{{\rm ST}, n,i}}.
	\end{align}
	Over the course of training, we compute the class distribution $\delta_n$ of each image through Eq.~\ref{eq7}. Then the class distribution $\delta_{pseudo}$ after each training iteration $k$ is updated with a momentum $\alpha\in[0,1]$: 
	\begin{align}
		{\delta}_{pseudo}^{(c)}|_{k} = \alpha {\delta}_{pseudo}^{(c)}|_{k-1}+(1-\alpha){\delta}_{n}^{(c)}.
	\end{align}
	Finally, we enforce the class distribution of cluster assignments, denoted by $r$ in Eq.~\ref{eq4}, to be close to ${\delta}_{pseudo}$:
	\begin{align}
		r={\delta}_{pseudo}, \quad h=\frac{1}{M}\mathbf{1}_M.
	\end{align}
	
	Our empirical results (please refer to Fig.~\ref{fig5}) demonstrate that the proposed distribution alignment technique effectively avoids the dominance of majority classes during training. Please refer to Sec.~\ref{sec4.3} for more discussions. 
	\begin{table*}[!ht]
		\centering
		\scalebox{0.65}{
			\setlength{\tabcolsep}{1.4mm}
			\begin{tabular}{c|ccccccccccccccccccc|c}
				\toprule[1.5pt]  
				\rowcolor{gray!20}
				Method   & \rotatebox{90}{\small road} & \rotatebox{90}{\small sideway} & \rotatebox{90}{\small building} & \rotatebox{90}{\small wall} & \rotatebox{90}{\small fence} & \rotatebox{90}{\small pole} & \rotatebox{90}{\small light} & \rotatebox{90}{\small sign} & \rotatebox{90}{\small vege} & \rotatebox{90}{\small terrace} & \rotatebox{90}{\small sky}  & \rotatebox{90}{\small person} & \rotatebox{90}{\small rider} & \rotatebox{90}{\small car} & \rotatebox{90}{\small truck} & \rotatebox{90}{\small bus}  & \rotatebox{90}{\small train} & \rotatebox{90}{\small motor} & \rotatebox{90}{\small bike} & mIoU \\ \midrule[1pt]
				AdaptSeg~\cite{tsai2018learning} & 86.5 & 25.9    & 79.8     & 22.1 & 20.0  & 23.6 & 33.1  & 21.8 & 81.8       & 25.9    & 75.9 & 57.3   & 26.2  & 76.3 & 29.8  & 32.1 & 7.2   & 29.5      & 32.5 & 41.4 \\
				CyCADA~\cite{hoffman2017cycada}   & 86.7 & 35.6    & 80.1     & 19.8 & 17.5  & 38.0 & 39.9  & 41.5 & 82.7       & 27.9    & 73.6 & 64.9   & 19.0  & 65.0 & 12.0  & 28.6 & 4.5   & 31.1      & 42.0 & 42.7 \\
				ADVENT~\cite{vu2019advent}   & 89.4 & 33.1    & 81.0     & 26.6 & 26.8  & 27.2 & 33.5  & 24.7 & 83.9       & 36.7    & 78.8 & 58.7   & 30.5  & 84.8 & 38.5  & 44.5 & 1.7   & 31.6      & 32.4 & 45.5 \\
				CBST~\cite{zou2018unsupervised}     & 91.8 & 53.5    & 80.5     & 32.7 & 21.0  & 34.0 & 28.9  & 20.4 & 83.9       & 34.2    & 80.9 & 53.1   & 24.0  & 82.7 & 30.3  & 35.9 & 16.0  & 25.9      & 42.8 & 45.9 \\
				FADA~\cite{wang2020classes}     & 92.5 & 47.5    & {\bf 85.1}     & 37.6 & 32.8  & 33.4 & 33.8  & 18.4 & 85.3       & 37.7    & {\bf 83.5} & 63.2   & {\bf 39.7}  & 87.5 & 32.9  & 47.8 & 1.6   & 34.9      & 39.5 & 49.2 \\
				CAG\_UDA~\cite{zhang2019category} & 90.4 & 51.6    & 83.8     & 34.2 & 27.8  & 38.4 & 25.3  & \textbf{48.4} & 85.4       & 38.2    & 78.1 & 58.6   & 34.6  & 84.7 & 21.9  & 42.7 & {\bf41.1}  & 29.3      & 37.2 & 50.2 \\
				FDA~\cite{yang2020fda}      & 92.5 & 53.3    & 82.4     & 26.5 & 27.6  & 36.4 & 40.6  & 38.9 & 82.3       & 39.8    & 78.0 & 62.6   & 34.4  & 84.9 & 34.1  & 53.1 & 16.9  & 27.7      & 46.4 & 50.5 \\
				PIT~\cite{lv2020cross}      & 87.5 & 43.4    & 78.8     & 31.2 & 30.2  & 36.3 & 39.3  & 42.0 & 79.2       & 37.1    & 79.3 & 65.4   & 37.5  & 83.2 & {\bf46.0}  & 45.6 & 25.7  & 23.5      & 49.9 & 50.6 \\
				IAST~\cite{mei2020instance}     & {\bf 93.8} & {\bf 57.8}    & {\bf 85.1}     & 39.5 & 26.7  & 26.2 & 43.1  & 34.7 & 84.9       & 32.9    & 88.0 & 62.6   & 29.0  & 87.3 & 39.2  & 49.6 & 23.2  & 34.7      & 39.6 & 51.5 \\
				ProDA~\cite{zhang2021prototypical}    & 91.5 & 52.4    & 82.9     & {42.0} & {\bf 35.7}  & 40.0 & 44.4  & 43.3 & {\bf 87.0}       & {\bf 43.8}    & 79.5 & 66.5   & 31.4  & 86.7 & 41.1  & 52.5 & 0.0   & 45.4      & {\bf53.8} & 53.7 \\  
				CPSL (ours)    &  91.7     &  52.9   &  83.6   &   {\bf43.0}   &    32.3   &  {\bf43.7}    &   {\bf51.3}    &    42.8        &   85.4      &   37.6   &   81.1     &   {\bf 69.5}    &  30.0    &   {\bf88.1}    &   44.1   & {\bf 59.9}      &     24.9      &  {\bf 47.2}  &  48.4  & {\bf 55.7}    \\
				\midrule[1pt]
				ProDA$+distill$ & 	87.8	&	56.0	&	79.7	&	\textbf{46.3}	&	\textbf{44.8}	&	45.6	& 	53.5	&	53.5	&	\textbf{88.6}	&	\textbf{45.2}	&	82.1	& 	70.7	&	\textbf{39.2}	&	88.8	&	45.5	&	59.4	&	1.0		&	48.9	&	\textbf{56.4 }	&	57.5\\
				CPSL$+distill$  	&	\textbf{92.3}	&	\textbf{59.9}	&	\textbf{84.9}	&	45.7	&	29.7	&	\textbf{52.8}	&	\textbf{61.5}	&	\textbf{59.5}	&	87.9	&	41.5	&	\textbf{85.0}	&	\textbf{73.0}	&	35.5	&	\textbf{90.4}	&	\textbf{48.7}	&	\textbf{73.9}	&	\textbf{26.3}	&	\textbf{53.8}	&	53.9	&	\textbf{60.8}	
				\\ \bottomrule[1.5pt]
		\end{tabular}}
		\vspace{-0.5em}
		\caption{Experimental results on the GTA5 $\to$ Cityscapes adaptation task. The top score is highlighted in {\bf bold} font.}
		\label{0tab1}
	\end{table*}
	
	\begin{table*}[!ht]
		\centering
		\scalebox{0.61}{
			\setlength{\tabcolsep}{1.85mm}
			\begin{tabular}{c|cccccccccccccccc|c|c}
				\toprule[1.5pt]  
				\rowcolor{gray!20}
				Method   & \rotatebox{90}{\small road} & \rotatebox{90}{\small sideway} & \rotatebox{90}{\small building} & \rotatebox{90}{\small wall} & \rotatebox{90}{\small fence} & \rotatebox{90}{\small pole} & \rotatebox{90}{\small light} & \rotatebox{90}{\small sign} & \rotatebox{90}{\small vege} & \rotatebox{90}{\small sky}  & \rotatebox{90}{\small person} & \rotatebox{90}{\small rider} & \rotatebox{90}{\small car}  & \rotatebox{90}{\small bus}  & \rotatebox{90}{\small motor} & \rotatebox{90}{\small bike} & mIoU$^{13}$ & mIoU$^{16}$ \\ \midrule[1pt]
				AdaptSeg~\cite{tsai2018learning} & 79.2 & 37.2    & 78.8     & -    & -     & -    & 9.9   & 10.5 & 78.2       & 80.5 & 53.5   & 19.6  & 67.0 & 29.5 & 21.6      & 31.3 & 45.9 & -    \\
				ADVENT~\cite{vu2019advent}   & 85.6 & 42.2    & 79.7     & 8.7  & 0.4   & 25.9 & 5.4   & 8.1  & 80.4       & 84.1 & 57.9   & 23.8  & 73.3 & 36.4 & 14.2      & 33.0 & 48.0 & 41.2 \\
				CBST~\cite{zou2018unsupervised}     & 68.0 & 29.9    & 76.3     & 10.8 & 1.4   & 33.9 & 22.8  & 29.5 & 77.6       & 78.3 & 60.6   & 28.3  & 81.6 & 23.5 & 18.8      & 39.8 & 48.9 & 42.6 \\
				CAG\_UDA\cite{zhang2019category} & 84.7 & 40.8    & 81.7     & 7.8  & 0.0   & 35.1 & 13.3  & 22.7 & 84.5       & 77.6 & 64.2   & 27.8  & 80.9 & 19.7 & 22.7      & 48.3 & 51.5 & 44.5 \\
				PIT~\cite{lv2020cross}      & 83.1 & 27.6    & 81.5     & 8.9  & 0.3   & 21.8 & 26.4  & {33.8}	& 76.4       & 78.8 & 64.2   & 27.6  & 79.6 & 31.2 & 31.0      & 31.3 & 51.8 & 44.0 \\
				FADA~\cite{wang2020classes}     & 84.5 & 40.1    & 83.1     & 4.8  & 0.0   & 34.3 & 20.1  & 27.2 & 84.8       & 84.0 & 53.5   & 22.6  & 85.4 & 43.7 & 26.8      & 27.8 & 52.5 & 45.2 \\
				FDA~\cite{yang2020fda}      & 79.3 & 35.0    & 73.2     & -    & -     & -    & 19.9  & 24.0 & 61.7       & 82.6 & 61.4   & \textbf{31.1 } & 83.9 & 40.8 & 38.4      & 51.1 & 52.5 & -    \\
				PyCDA~\cite{lian2019constructing}    & 75.5 & 30.9    & 83.3     & 20.8 & 0.7   & 32.7 & 27.3  & 33.5 & 84.7       & 85.0 & 64.1   & 25.4  & 85.0 & 45.2 & 21.2      & 32.0 & 	53.3  & 46.7 \\
				IAST~\cite{mei2020instance}     & 81.9 & 41.5    & 83.3     & 17.7 & \textbf{4.6}   & 32.3 & 30.9  & 28.8 & 83.4       & 85.0 & 65.5   & 30.8  & 86.5 & 38.2 & 33.1      & 52.7 & 57.0 & 49.8 \\
				SAC~\cite{araslanov2021self}      & \textbf{89.3} & \textbf{47.2 }   & \textbf{85.5 }    & 26.5 & 1.3   & \textbf{43.0} & 45.5  & 32.0 & \textbf{87.1 }      & \textbf{89.3}	& 63.6   & 25.4  & 86.9 & 35.6 & 30.4      & 53.0 & 59.3 & 52.6 \\
				ProDA~\cite{zhang2021prototypical}      & 87.1 &44.0   &83.2    & \textbf{26.9} & 0.7   & 42.0 & 45.8 & \textbf{34.2} & 86.7      & 81.3	& 68.4   & 22.1  & 87.7 & 50.0 & 31.4      & 38.6 &58.5  & 51.9\\
				CPSL (ours)    &   87.3   &    44.4     &    83.8      &   25.0   &   0.4    &   42.9   &   \textbf{47.5}    &   32.4   &     86.5       &    83.3  &    \textbf{69.6}    &   29.1    &   \textbf{89.4 }  &   \textbf{52.1}   &     \textbf{42.6 }     &  \textbf{ 54.1 }  &   \textbf{61.7}   &   \textbf{54.4 }  \\ 
				\midrule[1pt]
				ProDA$+distill$ & \textbf{87.8} 	&	\textbf{45.7}	&  84.6		& 	\textbf{37.1} 	&	\textbf{0.6} 	&	44.0	&	54.6	&	37.0	&	\textbf{88.1}	&	84.4	&	74.2	&	24.3	&	88.2	&	\textbf{51.1}	&	40.5	&	45.6	&  	62.0	&	55.5\\
				CPSL$+distill$ & 87.2 	&	43.9	&  \textbf{85.5}		& 	33.6 	&	0.3 	&	\textbf{47.7}	&	\textbf{57.4}	&	\textbf{37.2}	&	87.8	&	\textbf{88.5}	&	\textbf{79.0}	&	\textbf{32.0}	&	\textbf{90.6}	&	49.4	&	\textbf{50.8}	&	\textbf{59.8}	&  	\textbf{65.3}	&	\textbf{57.9}\\
				\bottomrule[1.5pt]
		\end{tabular}}
		\vspace{-0.5em}	
		\caption{Experimental results on the SYNTHIA $\to$ Cityscapes adaptation task. The top score is highlighted in {\bf bold} font.}
		\vspace{-1.0em}	
		\label{0tab2}
	\end{table*}

	\subsection{Loss Function}
	\label{sec3.5}
	As shown in Fig.~\ref{0fig1}, we employ momentum encoder to stabilize the self-labeling process. To further improve the model generalization ability on target domain and alleviate the bias inherited from source domain, following \cite{zhang2021prototypical,araslanov2021self}, we impose consistency regularization on the segmentation network. Specifically, we generate a weakly-augmented image $X_w$ and a strongly-augmented image $X_s$ from the same input image $X$, and pass $X_w$ through the momentum segmentation network $f'_{\rm SEG}$ to generate a probability map $P_w$, which is used to supervise the output $P_s$ of strongly-augmented image $X_s$ from $f_{\rm SEG}$. Then we enforce $P_w$ and $P_s$ to be consistent via:
	\begin{equation}
		{\small 	\begin{split}
				\mathcal{L}_{\rm REG}=\sum_{n=1}^{N_T} \sum_{i=1}^{H \times W}\left(\ell_{\rm KL}\left({P}_{w, n,i}, {P}_{s, n,i}\right)\right. 
				\left.+\ell_{\rm KL}\left({P}_{s, n,i}, {P}_{w, n,i}\right)\right),
			\end{split}
			\label{eq9}}
	\end{equation}
	where $\ell_{\rm KL}$ denotes the KL-divergence. $P_{s,n,i}$ and $P_{w,n,i}$ represent the $i$-th pixel of the segmentation probability maps $P_s$ and $P_w$ of image $X_n$, respectively. 
	
	The overall loss function is defined as:
	\begin{align}
		\mathcal{L}_{\rm TOTAL} = \mathcal{L}_{\rm SEG} + \lambda_1\mathcal{L}_{\rm SL} +\lambda_2\mathcal{L}_{\rm REG},
	\end{align}
	where $\lambda_1$ and $\lambda_2$ are trade-off parameters. $\mathcal{L}_{\rm SL}$ and $\mathcal{L}_{\rm REG}$ are complementary to each other. The former uses pixel-level cluster assignment $P_{\rm SL}$ to rectify the pseudo label $P_{\rm ST}$, which effectively dilutes the bias to source domain, while the latter improves model generalization ability by applying data augmentations on inputs and consistency regularization on outputs. 
	
	\begin{figure*}
		\centering \vspace{-1.5em}
		\includegraphics[scale=0.10]{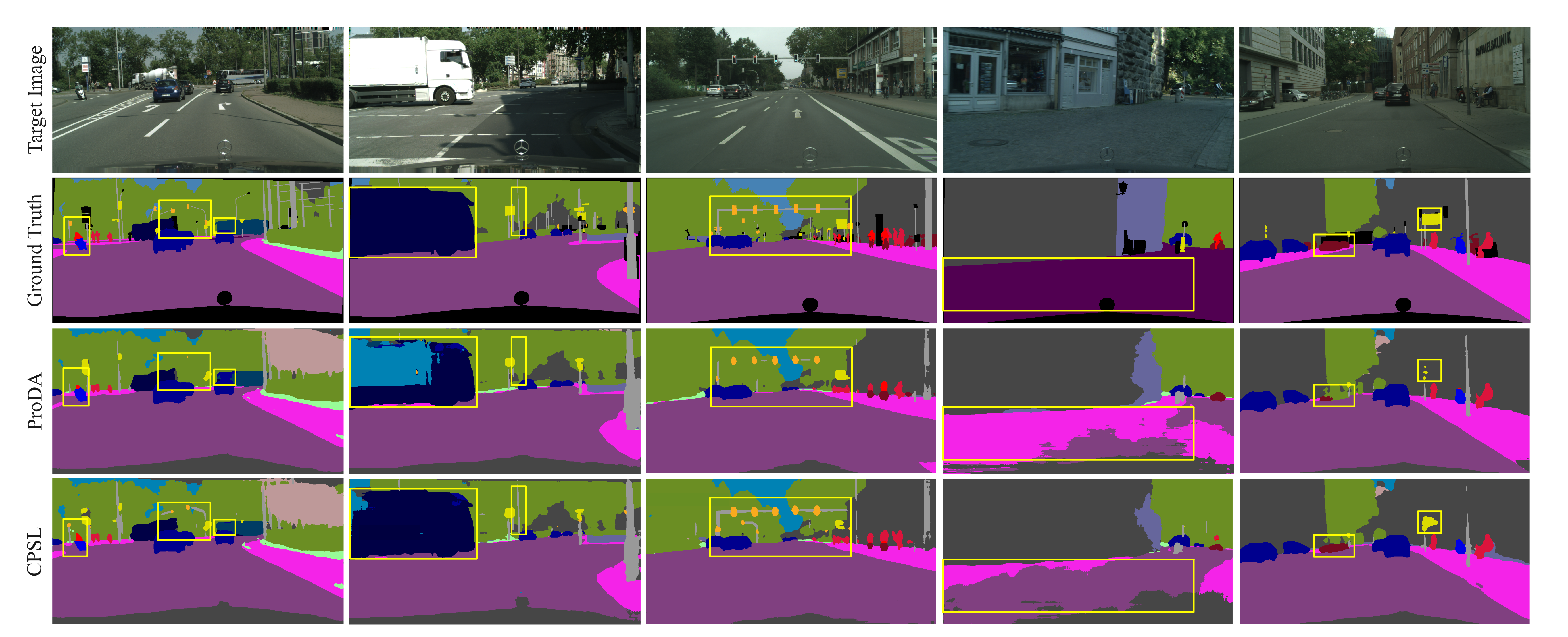}\\
		\vspace{-1.0em}
		\caption{Qualitative results of our method and ProDA~\cite{zhang2021prototypical} on the GTA5$\to$Cityscapes task.}
		\label{fig4}
		\vspace{-1.3em}
	\end{figure*}
	
	\section{Experiments}
	\subsection{Experimental Settings}
	\noindent{\bf Implementation Details.}
	We implement the segmentation model with DeepLabv2~\cite{chen2017deeplab} and employ  ResNet-101~\cite{he2016deep} as the backbone, which is pre-trained on ImageNet. The segmentation model is warmed up by applying adversarial training like~\cite{tsai2018learning}.
	The input images are randomly cropped to 896$\times$512, and the batch size is set as 4. We employ a series of data augmentations such as RandAugment~\cite{cubuk2020randaugment}, Cutout~\cite{devries2017improved}, CutMix~\cite{yun2019cutmix}, and add photometric noise, including color jitter, random blur, \etc. SGDM is used as the optimizer. The initial learning rate of segmentation model and self-labeling head are set to $10^{-4}$ and $5\times 10^{-4}$, which decay exponentially with power $0.9$. The weight decay and momentum are set to $2\times 10^{-4}$ and $0.9$, respectively. The trade-off parameters $\lambda_1$, $\lambda_2$ and the temperature parameters $\tau$, $\varepsilon$  are empirically set to 0.1, 5, 0.08, and 0.05, respectively. The length of memory bank is set to $65,536$ and we sample $512$ pixels per image for clustering ($M=512$), that is, there are 128 images in the memory bank. For the momentum networks, the momentum is set to 0.999. Our model is trained with four Tesla V100 GPUs on PyTorch. 
	
	\noindent{\bf Datasets.} Following ~\cite{zhang2019category,zhang2017curriculum,mei2020instance}, we adopt two synthetic datasets (GTA5~\cite{richter2016playing}, SYNTHIA~\cite{ros2016synthia}) and one real dataset (Cityscapes~\cite{cordts2016cityscapes}) in the experiments. The GTA5 dataset contains 24,966 images with resolution 1914$\times$1052. The corresponding dense annotations are generated by game engine. The SYNTHIA dataset contains 9,400 images of 1280$\times$760 pixels and it has 16 common categories with Cityscapes, which contains 2,975 training images and 500 validation images of resolution 2048$\times$1024. 
	
	\subsection{Comparisons with State-of-the-Arts}
	\label{sec4.2}
	We name the proposed method as Class-balanced Pixel-level Self-Labeling (CPSL). Following~\cite{zhang2021prototypical}, after the training converges, we also conduct two more knowledge distillation rounds to transfer the knowledge to a student model pre-trained in a self-supervised manner, and the resulting model is called “CPSL+distill”. 
	We compare our models with representative and state-of-the-art methods, which can be categorized to two main groups: adversarial training-based methods, including AdaptSeg~\cite{tsai2018learning}, CyCADA~\cite{hoffman2017cycada}, FADA~\cite{wang2020classes}, ADVENT~\cite{vu2019advent}, and self-training based methods, including CBST~\cite{zou2018domain}, IAST~\cite{mei2020instance}, CAG\_UDA~\cite{zhang2019category}, ProDA~\cite{zhang2021prototypical}, SAC~\cite{araslanov2021self}. Following previous works, the results on validation set are reported in terms of category-wise Intersection over Union (IoU) and mean IoU (mIoU).
	
	\vspace{0.1em}\noindent{\bf GTA5$\to$Cityscapes.}  The results on GTA5$\to$Cityscapes task are reported in Tab.~\ref{0tab1}. Our CPSL achieves the best IoU score on 7 out of 19 categories, and it achieves the highest mIoU score, outperforming the second best method ProDA~\cite{zhang2021prototypical} by a large margin of 2.0. This can be attributed to the exploration of inherent data distribution of target domain, which provides extra supervision for training. By applying knowledge distillation, there is a further performance gain of 5.1, achieving 60.8 mIoU, which is by far the new state-of-the-art. It is worth mentioning that our method performs especially well on long-tailed categories, such as ``pole'', ``light'', ``train'', and ``motor''. For example, ProDA fails on the small class ``train'' due to the difficulties in constructing prototypes for long-tailed categories. By applying distribution alignment, CPSL alleviates the class-imbalance problem, attaining 24.9 IoU on ``train'' without sacrificing the performance on other categories.
	\begin{table}[!t]
		\begin{minipage}[htbp]{\textwidth}
			\begin{minipage}[t]{0.23\textwidth}
				\makeatletter\def\@captype{table}\makeatother  
				\scalebox{0.8}{
					\begin{tabular}{c||c|c} 
						\toprule[1.0pt]  
						\rowcolor{gray!20}
						Configuration & mIoU & $\Delta$ \\ \hline\hline
						w/o SL        & 47.8 & -7.9     \\
						w/o CB        & 51.8 & -3.9     \\
						w/o ST        & 39.4 & -16.3    \\
						w/o Init      & 49.9 & -5.8     \\
						w/o Aug       & 54.2 & -1.5     \\
						w/o Mom       & 54.6 & -1.1     \\ \hline
						CPSL          & \textbf{55.7} & -       \\ \bottomrule[1.0pt]	
				\end{tabular}}
				\caption{Ablation studies on the key components of our proposed method.}
				\vspace{-1.5em}
				\label{0tab3}
			\end{minipage}
			\hspace{1.6em}
			\begin{minipage}[t]{0.20\textwidth}
				\makeatletter\def\@captype{table}\makeatother
				\scalebox{0.91}{
					\hspace{1.0em}\begin{tabular}{c|c}        
						\toprule[1.0pt]  
						\rowcolor{gray!20}
						{\small 	\# samples} & mIoU \\ \hline\hline
						64        & 54.9 \\
						128       &  55.3    \\
						256        &   55.5   \\
						512        & \textbf{55.7}   \\
						1024      & 54.3      \\
						2048       & 53.4    \\ \bottomrule[1.0pt]
				\end{tabular}}
				\caption{The influence of the number of samples per image on performance.}
				\vspace{-1.5em}
				\label{tab4}
			\end{minipage}
		\end{minipage}
	\end{table}
	
	\vspace{0.1em}\noindent{\bf SYNTHIA$\to$Cityscapes.} This adaptation task is more challenging than the previous one because of the large domain gap. The mIoUs over 13 classes (mIoU$^{13}$) and 16 classes (mIoU$^{16}$) are reported in Tab.~\ref{0tab2}. Our model still achieves significant improvements over competing methods on this task. Specifically, CPSL achieves the mIoU of 54.4 and 61.7 over 16 and 13 categories, surpassing the second best method SAC~\cite{araslanov2021self} by 1.8 and 2.4, respectively. This owes to the fact that CPSL reduces the label noise and calibrates the bias to source domain. The results are further improved to 57.9 and 65.3 in terms of mIoU after distillation. Among all the 16 categories, our method tops over six of them, especially on the hardest categories, such as ``light'', ``motorbike'', ``bike'', and so on. 
	
	\begin{figure}
		\vspace{-0.6em}\hspace{-0.8em}
		\includegraphics[scale=0.065]{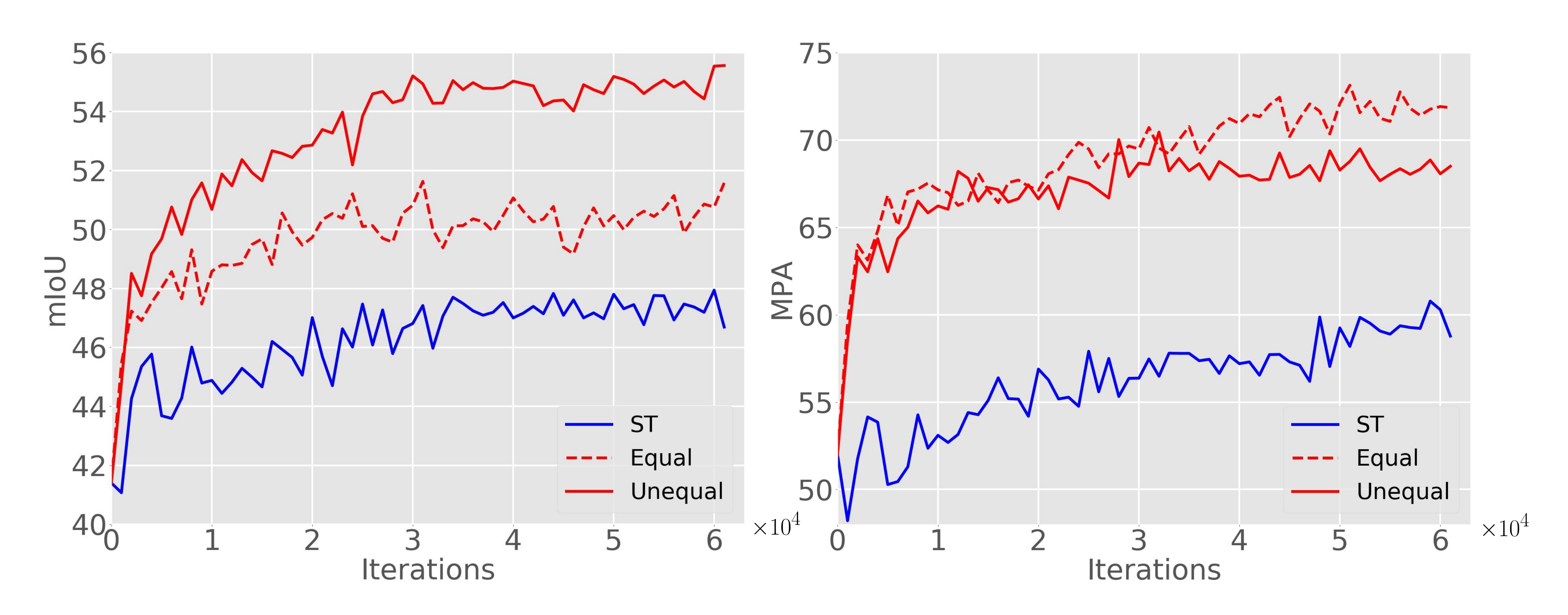}\\
		\vspace{-2.5em}
		\caption{The mIoU and mean pixel accuracy (MPA) scores evaluated on the validation set with equal/unequal partition constraint.}
		\label{fig7}
		\vspace{-1em}
	\end{figure}
	
	\begin{figure*}
		\centering \vspace{-1.5em}
		\includegraphics[scale=0.135]{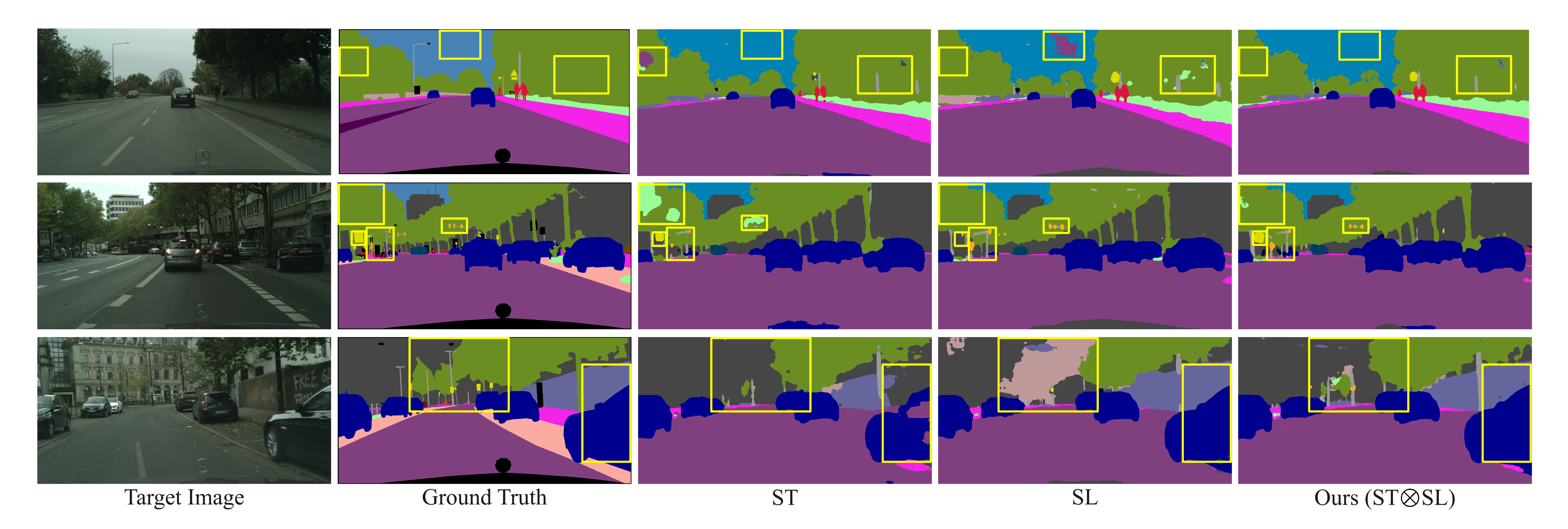}\\
		\vspace{-1.5em}
		\caption{The complementarity between label assignments produced by self-training (ST) and self-labeling (SL).}
		\label{fig3}
		\vspace{-1em}
	\end{figure*}
	
	\vspace{0.2em}	\noindent\textbf{Qualitative Results.}
	Fig.~\ref{fig4} shows the qualitative segmentation results of our method and ProDA~\cite{zhang2021prototypical} on GTA5$\to$Cityscapes task. As can be seen, our method improves the performance on long-tailed classes substantially, \eg~``pole'', ``light'', ``bus'', thanks to the class-balanced sampling and distribution alignment techniques. ProDA~\cite{zhang2021prototypical} does not perform well on these categories since it does not explicitly enforce class balance in training. 
	
	\begin{figure}
		\vspace{-0.5em}
		\centering
		\includegraphics[scale=0.110]{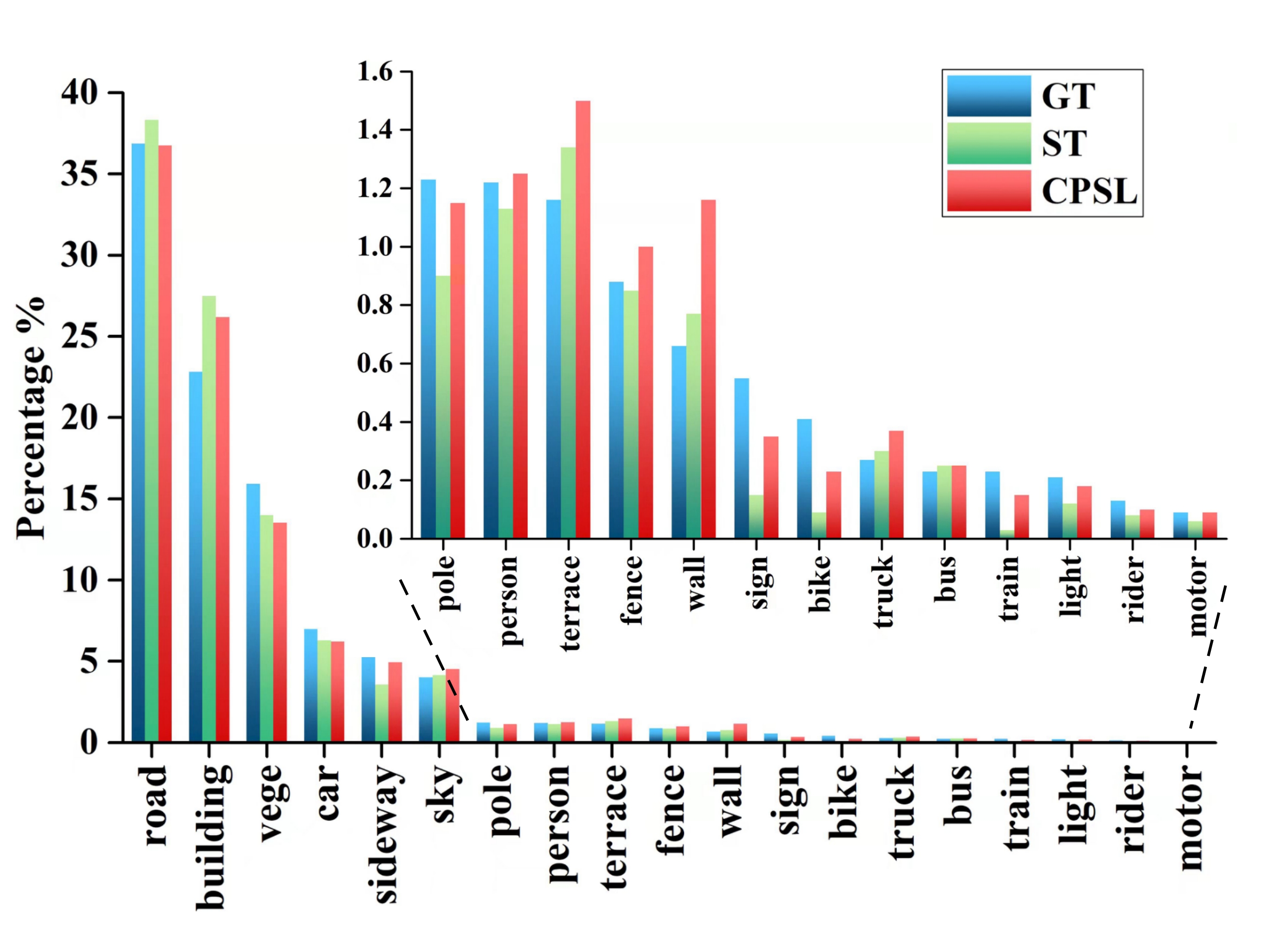}\\
		\vspace{-1.0em}
		\caption{The comparisons of class distributions on Cityscapes dataset. `GT' denotes the ground truth class distribution. `ST' and `CPSL' denote the class distributions of pseudo labels produced by self-training and class-balanced pixel-level self-labeling.}
		\label{fig5}
		\vspace{-1em}
	\end{figure}
	
	\subsection{Discussions}
	\label{sec4.3}
	\noindent\textbf{Ablation Study.} We conduct ablation studies on the GTA5$\to$Cityscapes task to investigate the role of each component in CPSL. For the convenience of expression, we abbreviate `self-labeling', `self-training', `class balance', `weight initialization', `data augmentation', and `momentum encoder' with `SL', `ST', `CB', `Init', `Aug', `Mom'. Tab.~\ref{0tab3} shows the corresponding results by switching off each component. We have the following observations.
	
	First, removing the SL component leads to a drop of 7.9 in mIoU, while disabling CB component leads to a drop of 3.9 in mIoU. This demonstrates they play key roles in improving the segmentation performance by exploring the intrinsic data structures of target domain images. Second, training without the pseudo labels produced by ST causes a significant drop of 16.3 in mIoU. This is not surprising because simultaneously updating network parameters and generating pseudo labels will lead to a degenerate solution~\cite{zhang2021prototypical,zhang2019category}. Third, randomly initializing the self-labeling head (w/o Init) results in a decline of 5.8 in mIoU, which is attributed to the mismatch between clustering and classification categories. Fourth, Aug and Mom components bring an improvement of 1.7 and 1.1 in mIoU. 
	
	\vspace{0.1em}	\noindent\textbf{Unequal Partition Constraint.} To further analyze the effect of unequal partition on class-imbalanced dataset, we plot the curves of mIoU and MPA scores with different partition constraints in Fig.~\ref{fig7}, where a huge gap can be observed in terms of mIoU. However, equal partition slightly outperforms unequal partition in terms of MPA. This is not surprising because many pixels belonging to large categories are assigned to small categories under the equal partition constraint, largely improving pixel accuracy of small classes without influencing much large classes. Thus the MPA score is improved. More details can be found in the \textit{supplemental files}.
	
	\vspace{0.1em}	\noindent\textbf{Self-Training (ST) vs. Self-Labeling (SL).} We explore the complementarity of label assignments produced by ST and SL, and visualize the results in Fig.~\ref{fig3}. One can draw a conclusion that the integration of ST and SL in our CPSL leads to better results than any one of them. Specifically, ST performs better on large categories which are easy to transfer, such as ``sky'' and ``building'', while SL has advantages on small categories such as ``light'' and ``pole''. Therefore, the pixels that are wrongly classified in one view will be corrected in another view.
	
	\vspace{0.1em}\noindent\textbf{The Effect of Distribution Alignment.} We compare the class distributions of labels produced by CPSL and conventional self-training~(ST). As illustrated in Fig.~\ref{fig5}, the results of ST mismatch heavily to ground truth (GT). Its predictions are biased towards majority categories, \eg `road' and `building', ignoring small categories such as `train', `sign' and `bike'. CPSL calibrates the bias and produces a class distribution closer to GT. This demonstrates that CPSL can capture the inherent class distribution of target domain and avoids gradual dominance of majority classes.
	
	\vspace{0.1em}\noindent\textbf{Parameter Sensitivity Analysis.} In Tab.~\ref{tab4}, we evaluate the segmentation performance on GTA5$\to$Cityscapes task with different number of samples per image. Our method is robust to this parameter within a wide range. More analyses can be found in \textit{supplemental materials}. 
	
	\vspace{0.1em}\noindent\textbf{Limitation.} Although the proposed CPSL alleviates the bias to source domain with the self-labeling assignment, it still relies on the self-training based pseudo labels, which may lead to confirmation bias. We consider to develop a fully clustering-based assignment method in future works.
	
	\section{Conclusion}
	We proposed a plug-and-play module, namely Class-balanced Pixel-level Self-Labeling (CPSL), which could be seamlessly incorporated into self-training pipelines to improve the domain adaptive semantic segmentation performance. Specifically, we conducted pixel-level clustering online and used the resulting cluster assignments to rectify pseudo labels. On one hand, the label noise was reduced and the bias to source domain was calibrated by exploring pixel-level intrinsic structures of target domain images. On the other hand, CPSL captured inherent class distribution of target domain, which effectively avoided gradual dominance of majority classes. Both the qualitative and quantitative analyses demonstrated that CPSL outperformed the existing state-of-the-arts by a large margin. In particular, it achieved great performance gains on long-tailed classes without sacrificing the performance on other categories.
	
\bibliographystyle{ieee_fullname}
\bibliography{egbib}

\begin{thebibliography}{10}\itemsep=-1pt

\bibitem{araslanov2021self}
Nikita Araslanov and Stefan Roth.
\newblock Self-supervised augmentation consistency for adapting semantic
  segmentation.
\newblock In {\em Proceedings of the IEEE/CVF Conference on Computer Vision and
  Pattern Recognition}, pages 15384--15394, 2021.

\bibitem{asano2020labelling}
Yuki Asano, Mandela Patrick, Christian Rupprecht, and Andrea Vedaldi.
\newblock Labelling unlabelled videos from scratch with multi-modal
  self-supervision.
\newblock {\em Advances in Neural Information Processing Systems},
  33:4660--4671, 2020.

\bibitem{asano2020self}
Y~M {Asano}, C {Rupprecht}, and A {Vedaldi}.
\newblock Self-labelling via simultaneous clustering and representation
  learning.
\newblock In {\em ICLR 2020 : Eighth International Conference on Learning
  Representations}, 2020.

\bibitem{caron2018deep}
Mathilde {Caron}, Piotr {Bojanowski}, Armand {Joulin}, and Matthijs {Douze}.
\newblock Deep clustering for unsupervised learning of visual features.
\newblock In {\em Proceedings of the European Conference on Computer Vision
  (ECCV)}, pages 132--149, 2018.

\bibitem{caron2020unsupervised}
Mathilde {Caron}, Ishan {Misra}, Julien {Mairal}, Priya {Goyal}, Piotr
  {Bojanowski}, and Armand {Joulin}.
\newblock Unsupervised learning of visual features by contrasting cluster
  assignments.
\newblock In {\em Thirty-fourth Conference on Neural Information Processing
  Systems (NeurIPS)}, volume~33, pages 9912--9924, 2020.

\bibitem{chang2019all}
Wei-Lun {Chang}, Hui-Po {Wang}, Wen-Hsiao {Peng}, and Wei-Chen {Chiu}.
\newblock All about structure: Adapting structural information across domains
  for boosting semantic segmentation.
\newblock In {\em 2019 IEEE/CVF Conference on Computer Vision and Pattern
  Recognition (CVPR)}, pages 1900--1909, 2019.

\bibitem{chen2019progressive}
Chaoqi {Chen}, Weiping {Xie}, Wenbing {Huang}, Yu {Rong}, Xinghao {Ding}, Yue
  {Huang}, Tingyang {Xu}, and Junzhou {Huang}.
\newblock Progressive feature alignment for unsupervised domain adaptation.
\newblock In {\em 2019 IEEE/CVF Conference on Computer Vision and Pattern
  Recognition (CVPR)}, pages 627--636, 2019.

\bibitem{chen2017deeplab}
Liang-Chieh Chen, George Papandreou, Iasonas Kokkinos, Kevin Murphy, and Alan~L
  Yuille.
\newblock Deeplab: Semantic image segmentation with deep convolutional nets,
  atrous convolution, and fully connected crfs.
\newblock {\em IEEE transactions on pattern analysis and machine intelligence},
  40(4):834--848, 2017.

\bibitem{chen2019crdoco}
Yun-Chun {Chen}, Yen-Yu {Lin}, Ming-Hsuan {Yang}, and Jia-Bin {Huang}.
\newblock Crdoco: Pixel-level domain transfer with cross-domain consistency.
\newblock In {\em 2019 IEEE/CVF Conference on Computer Vision and Pattern
  Recognition (CVPR)}, pages 1791--1800, 2019.

\bibitem{choi2019self}
Jaehoon {Choi}, Taekyung {Kim}, and Changick {Kim}.
\newblock Self-ensembling with gan-based data augmentation for domain
  adaptation in semantic segmentation.
\newblock In {\em 2019 IEEE/CVF International Conference on Computer Vision
  (ICCV)}, pages 6830--6840, 2019.

\bibitem{cordts2016cityscapes}
Marius Cordts, Mohamed Omran, Sebastian Ramos, Timo Rehfeld, Markus Enzweiler,
  Rodrigo Benenson, Uwe Franke, Stefan Roth, and Bernt Schiele.
\newblock The cityscapes dataset for semantic urban scene understanding.
\newblock In {\em Proceedings of the IEEE conference on computer vision and
  pattern recognition}, pages 3213--3223, 2016.

\bibitem{cubuk2020randaugment}
Ekin~D Cubuk, Barret Zoph, Jonathon Shlens, and Quoc~V Le.
\newblock Randaugment: Practical automated data augmentation with a reduced
  search space.
\newblock In {\em Proceedings of the IEEE/CVF Conference on Computer Vision and
  Pattern Recognition Workshops}, pages 702--703, 2020.

\bibitem{cuturi2013sinkhorn}
Marco {Cuturi}.
\newblock Sinkhorn distances: Lightspeed computation of optimal transport.
\newblock In {\em Advances in Neural Information Processing Systems 26},
  volume~26, pages 2292--2300, 2013.

\bibitem{devries2017improved}
Terrance DeVries and Graham~W Taylor.
\newblock Improved regularization of convolutional neural networks with cutout.
\newblock {\em arXiv preprint arXiv:1708.04552}, 2017.

\bibitem{fu2019dual}
Jun Fu, Jing Liu, Haijie Tian, Yong Li, Yongjun Bao, Zhiwei Fang, and Hanqing
  Lu.
\newblock Dual attention network for scene segmentation.
\newblock In {\em Proceedings of the IEEE/CVF Conference on Computer Vision and
  Pattern Recognition}, pages 3146--3154, 2019.

\bibitem{he2016deep}
Kaiming He, Xiangyu Zhang, Shaoqing Ren, and Jian Sun.
\newblock Deep residual learning for image recognition.
\newblock In {\em Proceedings of the IEEE conference on computer vision and
  pattern recognition}, pages 770--778, 2016.

\bibitem{hoffman2017cycada}
Judy {Hoffman}, Eric {Tzeng}, Taesung {Park}, Jun-Yan {Zhu}, Phillip {Isola},
  Kate {Saenko}, Alexei~A. {Efros}, and Trevor {Darrell}.
\newblock Cycada: Cycle-consistent adversarial domain adaptation.
\newblock In {\em International Conference on Machine Learning}, pages
  1989--1998, 2017.

\bibitem{hoffman2016fcns}
Judy {Hoffman}, Dequan {Wang}, Fisher {Yu}, and Trevor {Darrell}.
\newblock Fcns in the wild: Pixel-level adversarial and constraint-based
  adaptation.
\newblock {\em arXiv preprint arXiv:1612.02649}, 2016.

\bibitem{hong2018conditional}
Weixiang {Hong}, Zhenzhen {Wang}, Ming {Yang}, and Junsong {Yuan}.
\newblock Conditional generative adversarial network for structured domain
  adaptation.
\newblock In {\em 2018 IEEE/CVF Conference on Computer Vision and Pattern
  Recognition}, pages 1335--1344, 2018.

\bibitem{hu2018squeeze}
Jie Hu, Li Shen, and Gang Sun.
\newblock Squeeze-and-excitation networks.
\newblock In {\em Proceedings of the IEEE conference on computer vision and
  pattern recognition}, pages 7132--7141, 2018.

\bibitem{huang2019unsupervised}
Jiabo {Huang}, Qi {Dong}, Shaogang {Gong}, and Xiatian {Zhu}.
\newblock Unsupervised deep learning by neighbourhood discovery.
\newblock In {\em International Conference on Machine Learning}, pages
  2849--2858, 2019.

\bibitem{huang2019ccnet}
Zilong Huang, Xinggang Wang, Lichao Huang, Chang Huang, Yunchao Wei, and Wenyu
  Liu.
\newblock Ccnet: Criss-cross attention for semantic segmentation.
\newblock In {\em Proceedings of the IEEE/CVF International Conference on
  Computer Vision}, pages 603--612, 2019.

\bibitem{kim2020learning}
Myeongjin {Kim} and Hyeran {Byun}.
\newblock Learning texture invariant representation for domain adaptation of
  semantic segmentation.
\newblock In {\em 2020 IEEE/CVF Conference on Computer Vision and Pattern
  Recognition (CVPR)}, pages 12975--12984, 2020.

\bibitem{li2021t}
Ruihuang Li, Xu Jia, Jianzhong He, Shuaijun Chen, and Qinghua Hu.
\newblock T-svdnet: Exploring high-order prototypical correlations for
  multi-source domain adaptation.
\newblock In {\em Proceedings of the IEEE/CVF International Conference on
  Computer Vision}, pages 9991--10000, 2021.

\bibitem{li2019bidirectional}
Yunsheng {Li}, Lu {Yuan}, and Nuno {Vasconcelos}.
\newblock Bidirectional learning for domain adaptation of semantic
  segmentation.
\newblock In {\em 2019 IEEE/CVF Conference on Computer Vision and Pattern
  Recognition (CVPR)}, pages 6936--6945, 2019.

\bibitem{lian2019constructing}
Qing {Lian}, Lixin {Duan}, Fengmao {Lv}, and Boqing {Gong}.
\newblock Constructing self-motivated pyramid curriculums for cross-domain
  semantic segmentation: A non-adversarial approach.
\newblock In {\em 2019 IEEE/CVF International Conference on Computer Vision
  (ICCV)}, pages 6757--6766, 2019.

\bibitem{lin2017refinenet}
Guosheng Lin, Anton Milan, Chunhua Shen, and Ian Reid.
\newblock Refinenet: Multi-path refinement networks for high-resolution
  semantic segmentation.
\newblock In {\em Proceedings of the IEEE conference on computer vision and
  pattern recognition}, pages 1925--1934, 2017.

\bibitem{long2015fully}
Jonathan Long, Evan Shelhamer, and Trevor Darrell.
\newblock Fully convolutional networks for semantic segmentation.
\newblock In {\em Proceedings of the IEEE conference on computer vision and
  pattern recognition}, pages 3431--3440, 2015.

\bibitem{luo2019taking}
Yawei {Luo}, Liang {Zheng}, Tao {Guan}, Junqing {Yu}, and Yi {Yang}.
\newblock Taking a closer look at domain shift: Category-level adversaries for
  semantics consistent domain adaptation.
\newblock In {\em 2019 IEEE/CVF Conference on Computer Vision and Pattern
  Recognition (CVPR)}, pages 2507--2516, 2019.

\bibitem{lv2020cross}
Fengmao {Lv}, Tao {Liang}, Xiang {Chen}, and Guosheng {Lin}.
\newblock Cross-domain semantic segmentation via domain-invariant interactive
  relation transfer.
\newblock In {\em 2020 IEEE/CVF Conference on Computer Vision and Pattern
  Recognition (CVPR)}, pages 4334--4343, 2020.

\bibitem{mei2020instance}
Ke {Mei}, Chuang {Zhu}, Jiaqi {Zou}, and Shanghang {Zhang}.
\newblock Instance adaptive self-training for unsupervised domain adaptation.
\newblock In {\em European Conference on Computer Vision}, pages 415--430,
  2020.

\bibitem{melas2021pixmatch}
Luke Melas-Kyriazi and Arjun~K Manrai.
\newblock Pixmatch: Unsupervised domain adaptation via pixelwise consistency
  training.
\newblock In {\em Proceedings of the IEEE/CVF Conference on Computer Vision and
  Pattern Recognition}, pages 12435--12445, 2021.

\bibitem{murez2018image}
Zak {Murez}, Soheil {Kolouri}, David {Kriegman}, Ravi {Ramamoorthi}, and
  Kyungnam {Kim}.
\newblock Image to image translation for domain adaptation.
\newblock In {\em 2018 IEEE/CVF Conference on Computer Vision and Pattern
  Recognition}, pages 4500--4509, 2018.

\bibitem{pan2020unsupervised}
Fei {Pan}, Inkyu {Shin}, Francois {Rameau}, Seokju {Lee}, and In~So {Kweon}.
\newblock Unsupervised intra-domain adaptation for semantic segmentation
  through self-supervision.
\newblock In {\em 2020 IEEE/CVF Conference on Computer Vision and Pattern
  Recognition (CVPR)}, pages 3764--3773, 2020.

\bibitem{qi2018low}
Hang Qi, Matthew Brown, and David~G Lowe.
\newblock Low-shot learning with imprinted weights.
\newblock In {\em Proceedings of the IEEE conference on computer vision and
  pattern recognition}, pages 5822--5830, 2018.

\bibitem{richter2016playing}
Stephan~R Richter, Vibhav Vineet, Stefan Roth, and Vladlen Koltun.
\newblock Playing for data: Ground truth from computer games.
\newblock In {\em European conference on computer vision}, pages 102--118.
  Springer, 2016.

\bibitem{ros2016synthia}
German Ros, Laura Sellart, Joanna Materzynska, David Vazquez, and Antonio~M
  Lopez.
\newblock The synthia dataset: A large collection of synthetic images for
  semantic segmentation of urban scenes.
\newblock In {\em Proceedings of the IEEE conference on computer vision and
  pattern recognition}, pages 3234--3243, 2016.

\bibitem{snell2017prototypical}
Jake {Snell}, Kevin {Swersky}, and Richard~S. {Zemel}.
\newblock Prototypical networks for few-shot learning.
\newblock In {\em Advances in Neural Information Processing Systems},
  volume~30, pages 4077--4087, 2017.

\bibitem{tsai2018learning}
Yi-Hsuan {Tsai}, Wei-Chih {Hung}, Samuel {Schulter}, Kihyuk {Sohn}, Ming-Hsuan
  {Yang}, and Manmohan {Chandraker}.
\newblock Learning to adapt structured output space for semantic segmentation.
\newblock In {\em 2018 IEEE/CVF Conference on Computer Vision and Pattern
  Recognition}, pages 7472--7481, 2018.

\bibitem{tsai2019domain}
Yi-Hsuan {Tsai}, Kihyuk {Sohn}, Samuel {Schulter}, and Manmohan {Chandraker}.
\newblock Domain adaptation for structured output via discriminative patch
  representations.
\newblock In {\em 2019 IEEE/CVF International Conference on Computer Vision
  (ICCV)}, pages 1456--1465, 2019.

\bibitem{vu2019advent}
Tuan-Hung {Vu}, Himalaya {Jain}, Maxime {Bucher}, Matthieu {Cord}, and Patrick
  {Perez}.
\newblock Advent: Adversarial entropy minimization for domain adaptation in
  semantic segmentation.
\newblock In {\em 2019 IEEE/CVF Conference on Computer Vision and Pattern
  Recognition (CVPR)}, pages 2517--2526, 2019.

\bibitem{wang2020classes}
Haoran {Wang}, Tong {Shen}, Wei {Zhang}, Lingyu {Duan}, and Tao {Mei}.
\newblock Classes matter: A fine-grained adversarial approach to cross-domain
  semantic segmentation.
\newblock In {\em ECCV (14)}, pages 642--659, 2020.

\bibitem{wang2021exploring}
Wenguan Wang, Tianfei Zhou, Fisher Yu, Jifeng Dai, Ender Konukoglu, and Luc
  Van~Gool.
\newblock Exploring cross-image pixel contrast for semantic segmentation.
\newblock In {\em Proceedings of the IEEE/CVF International Conference on
  Computer Vision}, pages 7303--7313, 2021.

\bibitem{wu2018dcan}
Zuxuan {Wu}, Xintong {Han}, Yen-Liang {Lin}, Mustafa~Gökhan {Uzunbas}, Tom
  {Goldstein}, Ser~Nam {Lim}, and Larry~S. {Davis}.
\newblock Dcan: Dual channel-wise alignment networks for unsupervised scene
  adaptation.
\newblock In {\em Proceedings of the European Conference on Computer Vision
  (ECCV)}, pages 518--534, 2018.

\bibitem{xie2016unsupervised}
Junyuan {Xie}, Ross {Girshick}, and Ali {Farhadi}.
\newblock Unsupervised deep embedding for clustering analysis.
\newblock In {\em ICML'16 Proceedings of the 33rd International Conference on
  International Conference on Machine Learning - Volume 48}, pages 478--487,
  2016.

\bibitem{yan2020clusterfit}
Xueting {Yan}, Ishan {Misra}, Abhinav {Gupta}, Deepti {Ghadiyaram}, and Dhruv
  {Mahajan}.
\newblock Clusterfit: Improving generalization of visual representations.
\newblock In {\em 2020 IEEE/CVF Conference on Computer Vision and Pattern
  Recognition (CVPR)}, pages 6509--6518, 2020.

\bibitem{yang2016joint}
Jianwei {Yang}, Devi {Parikh}, and Dhruv {Batra}.
\newblock Joint unsupervised learning of deep representations and image
  clusters.
\newblock In {\em 2016 IEEE Conference on Computer Vision and Pattern
  Recognition (CVPR)}, pages 5147--5156, 2016.

\bibitem{yang2020fda}
Yanchao Yang and Stefano Soatto.
\newblock Fda: Fourier domain adaptation for semantic segmentation.
\newblock In {\em Proceedings of the IEEE/CVF Conference on Computer Vision and
  Pattern Recognition}, pages 4085--4095, 2020.

\bibitem{yun2019cutmix}
Sangdoo Yun, Dongyoon Han, Seong~Joon Oh, Sanghyuk Chun, Junsuk Choe, and
  Youngjoon Yoo.
\newblock Cutmix: Regularization strategy to train strong classifiers with
  localizable features.
\newblock In {\em Proceedings of the IEEE/CVF International Conference on
  Computer Vision}, pages 6023--6032, 2019.

\bibitem{zhang2021prototypical}
Pan Zhang, Bo Zhang, Ting Zhang, Dong Chen, Yong Wang, and Fang Wen.
\newblock Prototypical pseudo label denoising and target structure learning for
  domain adaptive semantic segmentation.

\bibitem{zhang2019category}
Qiming Zhang, Jing Zhang, Wei Liu, and Dacheng Tao.
\newblock Category anchor-guided unsupervised domain adaptation for semantic
  segmentation.
\newblock {\em arXiv preprint arXiv:1910.13049}, 2019.

\bibitem{zhang2017curriculum}
Yang {Zhang}, Philip {David}, and Boqing {Gong}.
\newblock Curriculum domain adaptation for semantic segmentation of urban
  scenes.
\newblock In {\em 2017 IEEE International Conference on Computer Vision
  (ICCV)}, pages 2039--2049, 2017.

\bibitem{zhao2017pyramid}
Hengshuang Zhao, Jianping Shi, Xiaojuan Qi, Xiaogang Wang, and Jiaya Jia.
\newblock Pyramid scene parsing network.
\newblock In {\em Proceedings of the IEEE conference on computer vision and
  pattern recognition}, pages 2881--2890, 2017.

\bibitem{zhuang2019local}
Chengxu {Zhuang}, Alex {Zhai}, and Daniel {Yamins}.
\newblock Local aggregation for unsupervised learning of visual embeddings.
\newblock In {\em 2019 IEEE/CVF International Conference on Computer Vision
  (ICCV)}, pages 6002--6012, 2019.

\bibitem{zou2018domain}
Yang Zou, Zhiding Yu, BVK Kumar, and Jinsong Wang.
\newblock Unsupervised domain adaptation for semantic segmentation via
  class-balanced self-training.
\newblock In {\em Proceedings of the European conference on computer vision
  (ECCV)}, pages 289--305, 2018.

\bibitem{zou2018unsupervised}
Yang {Zou}, Zhiding {Yu}, B.~V. K.~Vijaya {Kumar}, and Jinsong {Wang}.
\newblock Unsupervised domain adaptation for semantic segmentation via
  class-balanced self-training.
\newblock In {\em Proceedings of the European Conference on Computer Vision
  (ECCV)}, pages 297--313, 2018.

\bibitem{zou2019confidence}
Yang {Zou}, Zhiding {Yu}, Xiaofeng {Liu}, B.~V. K.~Vijaya {Kumar}, and Jinsong
  {Wang}.
\newblock Confidence regularized self-training.
\newblock In {\em 2019 IEEE/CVF International Conference on Computer Vision
  (ICCV)}, pages 5982--5991, 2019.

\bibitem{bautista2016cliquecnn}
Miguel Ángel {Bautista}, Artsiom {Sanakoyeu}, Ekaterina {Tikhoncheva}, and
  Björn {Ommer}.
\newblock Cliquecnn: Deep unsupervised exemplar learning.
\newblock In {\em Advances in Neural Information Processing Systems},
  volume~29, pages 3846--3854, 2016.

\end{thebibliography}

\clearpage
\section{Supplemental Materials}
	In this supplemental file, we provide the following materials:
	\begin{itemize}
	    \item[$\bullet$] The training procedure of CPSL;
		\item[$\bullet$] The definition of mean pixel accuracy (MPA) (referring to Sec4.3-Unequal partition constraint in the main paper);
		\item[$\bullet$] Ablation studies in terms of per-category IoU (referring to Sec4.3-Ablation study in the main paper);
		\item[$\bullet$] Comparisons on the training process on the GTA5$\to$Cityscapes task;
		\item[$\bullet$] More parameter sensitivity analyses (referring to Sec4.3-Parameter sensitivity analysis in the main paper);
		\item[$\bullet$] More qualitative results (referring to Sec4.2-Qualitative results  in the main paper).
	\end{itemize}
\subsection{Algorithm}
The training procedure of our CPSL is summarized in Algorithm.~\ref{alg2}. For detailed equations and loss functions, please refer to our main paper.
\begin{algorithm}[!h]
	\SetKwInOut{Input}{Input}\SetKwInOut{Output}{Output}
	\caption{Training Procedure of CPSL}\label{alg2}
	\Input{Training data $\mathcal{D}_{S}=\{(X^s_n,Y^s_n)\}^{N_S}_{n=1}$ and $\mathcal{D}_{T}=\{X^t_n\}^{N_T}_{n=1}$\;}
	\Output{The output model $f_{\rm SEG}$ \;}
	{Generate soft pseudo labels $P_{\rm ST}$ with the warmed-up model;}\\
	{Initialize the weight of $f_{\rm SL}$ and $f'_{\rm SL}$ with the prototypes $[\bar{\bf{z}}_1,\cdots,\bar{\bf{z}}_C]$ for each category computed by Eq.~7;}\\
	\For{$i=1$ to $max\_epochs$}
	{	\For{$n=1$ to $N_S$}
	    {   Get source image $X^s_n$;\\
	        Train the model $f_{\rm SEG}$ using loss $\mathcal{L}^s_{\rm SEG}$;\\
	        \vspace{1.5em}
	        Get target image $X^t_n$;\\
	        Extract features from $X^t_n$ to obtain ${Z}\in \mathbb{R}^{H\times W\times D}$ and normalize it with ${{ z}_i}=\frac{{ z}_i}{||{ z}_i||_2}$;\\
	        Sample a group of pixels $\hat{Z}=[{ z}_1,\cdots,{ z}_M]$ from $Z$ randomly;\\
	        Augment the features $\hat{Z}$ with a memory bank $\mathcal{M}$ and obtain $Z_{aug}=[\hat{Z};\mathcal{M}]$;\\
	        \For{$k=1$ to sinkhorn\_iterations}
	            {$Q^*_{aug} = {\diag}(\alpha)\exp(\frac{f_{\rm SL}(Z_{aug})}{\varepsilon})\diag(\beta)$;
	            }
	       Compute the self-labeling loss $\mathcal{L}_{\rm SL}$ through Eq.~5 using the cluster assignment of current batch $Q_{cur}$;\\
	       Train the self-labeling head $f_{\rm SL}$ using loss $\mathcal{L}_{\rm SL}$. \\
	       \vspace{1.5em}
	       Update the momentum self-labeling head $f'_{\rm SL}$ in an EMA manner;\\
	       Pass $X^t_n$ through $f'_{\rm SEG}$ and $f'_{\rm SL}$ to obtain self-labeling assignment $P_{\rm SL}$;\\
	       Use $P_{\rm SL}$ to rectify $P_{\rm ST}$ and obtain the rectified pseudo labels $\hat{Y}_n^t$ through Eq.~1;\\
	       Update $f_{\rm SEG}$ using loss $\mathcal{L}^t_{\rm SEG}$;\\
	       Update the momentum segmentation model $f'_{\rm SEG}$ in an EMA manner. 
	    }
	}
\end{algorithm}
		\begin{figure*}
		\hspace{-1.0em}\centering 
		\includegraphics[scale=0.10]{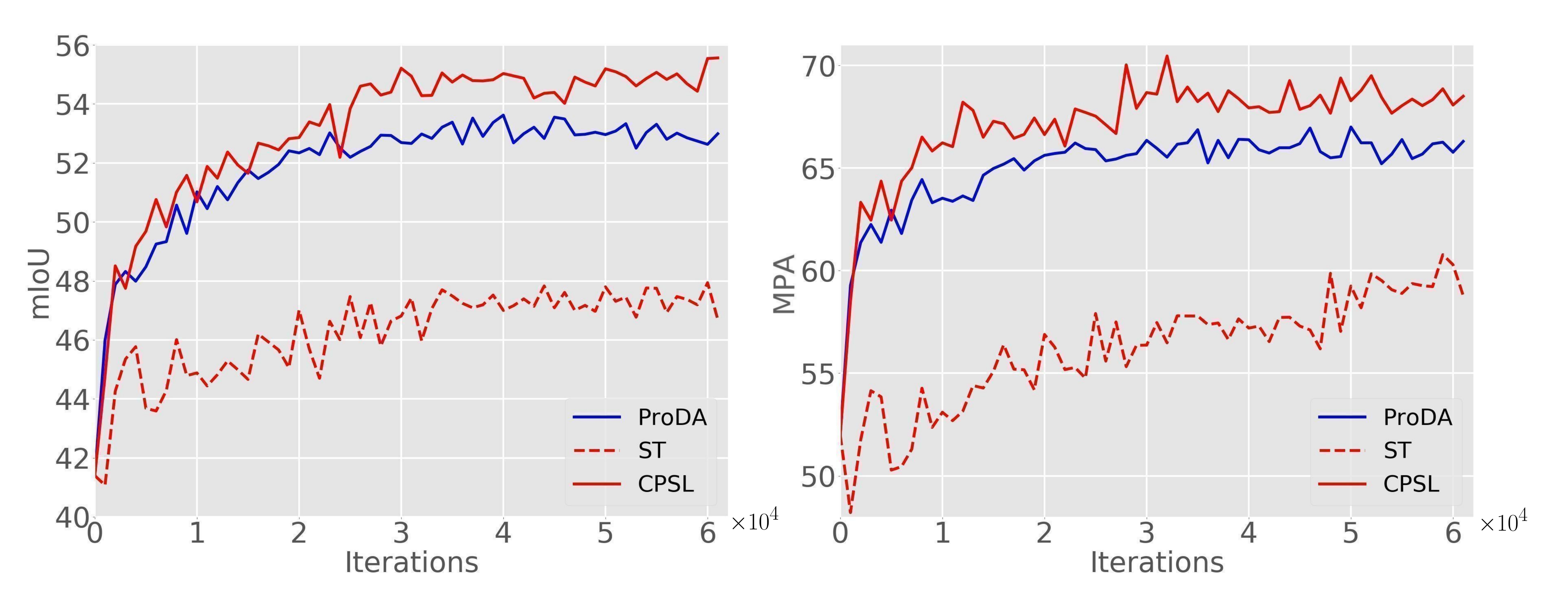}\\
		\vspace{-0.5em}
		\caption{The mIoU (left) and MPA (right) scores evaluated on the validation set during the training.}
		\label{loss}
	\end{figure*}
	
	\begin{table*}[!t]
		\centering
		\scalebox{0.74}{
			\setlength{\tabcolsep}{1.7mm}
			\begin{tabular}{c|ccccccccccccccccccc|cc}
				\toprule[1.5pt] 
				\rowcolor{gray!20}
				Method & {\rotatebox{90}{\small road}} & {\rotatebox{90}{\small sideway}} & {\rotatebox{90}{\small building}} & {\rotatebox{90}{\small wall}} & {\rotatebox{90}{\small fence}} & {\rotatebox{90}{\small pole}} & \rotatebox{90}{\small light} & \rotatebox{90}{\small sign} & \rotatebox{90}{\small vege} & \rotatebox{90}{\small terrace} & \rotatebox{90}{\small sky}  & \rotatebox{90}{\small person} & \rotatebox{90}{\small rider} & \rotatebox{90}{\small car} & \rotatebox{90}{\small truck} & \rotatebox{90}{\small bus}  & \rotatebox{90}{\small train} & \rotatebox{90}{\small motor} & \rotatebox{90}{\small bike}  & mIoU    & $\Delta$ \\ \midrule[1pt]
				w/o SL    & 91.9  & 56.3  & 82.9  & 35.9  & 30.2  & 37.5  & 37.4  & 32.9  & 85.3  &  39.2  &  77.8   &  51.2  & 18.6   &  84.7  & 37.8   & 44.6   & 1.0   &  20.2  &  42.7  & 47.8  &  -7.9  \\
				w/o ST&      82.4     &      39.0     & 70.5       &   30.5  & 16.0  & 24.1  & 39.6  & 37.0  & 77.8  & 24.2  & 78.7  & 28.5  &  18.7  &  75.7  &  9.2  &  36.1  &  4.1  &  22.9  & 36.5   &  39.4  & -16.3    \\
				w/o CB           &      91.7        &       51.3       &       { 84.0}        &  33.9  & 24.3  & 42.5 & 43.3  & \bf{49.0}  & 81.5  & 29.1  & 75.8  & 67.0  &  28.5  &  87.7  &  34.3  &  \bf{63.3}  & 20.1   & 36.0   &  40.5  &   51.8  & -3.9  \\
				w/o Init            &  89.6    & {\bf56.1}         &     80.0          &  40.3 & {\bf 36.7}  & 43.7  & 45.9  & 39.6 & 86.2  & 39.8  &  {\bf81.9}  &  66.7  &  24.8  & {\bf89.0}   & {\bf45.4}   &  50.8  & 0.0   &  31.4   & 9.3 & 49.9  & -5.8  \\
				w/o Aug            & 90.6             & 45.5           &   83.8    &  41.4   & 33.0  & 44.3  & {\bf 52.0}  & 42.0  &  {\bf 86.4}  &  40.2   & 81.6   &  68.4  &  28.9  &  88.0  &  42.8  &  58.5  & 14.9 &40.0 &47.1 & 54.2 & -1.5    \\
				w/o Mom            & {\bf 92.6}            & {53.7}           &   {\bf84.1}    &  41.7  & { 36.6}  & {\bf 44.8}  &  50.6 & 41.7  & 86.2  & {\bf 40.5}  &  79.6  & 68.2  &  26.6  & 87.4   &  37.4  &  55.9  &  19.3  &  43.1  & 47.5   &  54.6  & -1.1        \\
				CPSL            &    91.7   & 52.9    & 83.6      &  {\bf 43.0}  &  32.3  & 43.7  &  51.3  & 42.8  & 85.4  & 37.6  &  81.1  & {\bf69.5}  &  {\bf30.0}  &  88.1  &  44.1  &  {59.9}  &  {\bf24.9}  &  {\bf47.2}  &  {\bf48.4}  &   {\bf55.7}  & -    \\ \bottomrule[1.5pt] 
		\end{tabular}}
		\vspace{-0.5em}	
		\caption{Ablation studies on the key components of CPSL in terms of per-category IoU. The top score is highlighted in {\bf bold} font.} 
		\label{ablation}
		\vspace{-1.0em}	
	\end{table*}
	\subsection{Mean pixel accuracy (MPA)}
	Denoting by $C$ the number of classes, by $p_{ij}$ the number of pixels which belong to the $i$-th class but are wrongly classified into the $j$-th class, and by $p_{ii}$ the number of pixels which belong to the $i$-th class and are accurately classified into the $i$-th class, the pixel accuracy (PA) of the $i$-th class is defined as:
	\begin{equation}
		PA=\frac{p_{ii}}{\sum_{j=1}^{C}p_{ij}}.
	\end{equation}
	Then the mean pixel accuracy (MPA) is defined as:
	\begin{equation}
		MPA=\frac{1}{C}\sum_{i=1}^{C}\frac{p_{ii}}{\sum_{j=1}^{C}p_{ij}}. 
	\end{equation}
	As discussed in Sec.~4.3 of our manuscript, under the constraint of equal partition, many pixels belonging to large categories are assigned to small categories, largely improving the pixel accuracy of small classes. However, this constraint has very small influences on large categories because these categories contain a great number of pixels. Therefore, the MPA is improved. 
	
	\subsection{Ablation study}
	We only reported the mIoU scores in Tab.3 of the main paper. Here we present in Tab.~\ref{ablation} the per-class IoU scores of ablation studies. Note that ``w/o CB'' denotes that we do not employ the class-balanced sampling techniques, and constrain that $Q$ should induce an equipartition of data rather than an unequal partition. One can see that this leads to a degradation of 3.9 in terms of mIoU, demonstrating that the equal partition is not reasonable when the class distribution of data is highly imbalanced.

	\subsection{Training process of CPSL and ProDA}
	To further highlight the improvement of CPSL during training, we plot the curves of mIoU and MPA scores on the GTA5$\to$Cityscapes task in Fig.~\ref{loss}. A large performance improvement of CPSL over ProDA can be observed in terms of both mIoU and MPA.

	\subsection{Parameter analysis}
	Tab.~\ref{tab1} and Tab.~\ref{tab2} show the segmentation results by using different self-labeling loss weight $\lambda_1$ and consistency regularization loss weight $\lambda_2$, respectively. One can see that our method is insensitive to these two parameters. Tab.~\ref{tab3} shows the effect of temperature $\tau$. We employ the cluster assignment $P_{\rm SL}$ as a weight map to online modulate the softmax probability of pseudo labels $P_{\rm ST}$, where the temperature $\tau$ controls the modulation intensity. When $\tau \to 0$, the modulation intensity increases so that the rectified pseudo labels $\hat{Y}^t$ will rely heavily on $P_{\rm SL}$. When $\tau \to \infty$, the modulation intensity decreases so that the rectified pseudo labels $\hat{Y}^t$ will rely heavily on $P_{\rm ST}$.
	\begin{table}[!h]
		\centering
		\begin{tabular}{ccccc}
			\hline
			$\lambda_1$ & 0    & 0.01 & 0.1  & 0.5  \\ \hline
			mIoU                      & 51.4 & 54.2 & 55.7 & 54.9 \\ \hline
		\end{tabular}
		\caption{The influence of parameter $\lambda_1$.}
		\label{tab1}
	\end{table}
	
	\begin{table}[!h]
		\centering
		\begin{tabular}{cccccc}
			\hline
			$\lambda_2$ & 1    & 5    & 10   & 20   & 30   \\ \hline
			mIoU       & 55.5 & 55.7 & 55.2 & 54.7 & 54.4 \\ \hline
		\end{tabular}
		\caption{The influence of parameter $\lambda_2$.}
		\label{tab2}
	\end{table}

	\begin{table}[!h]
		\centering
		\begin{tabular}{ccccc}
			\hline
			$\tau$ &0.05 & 0.08 & 0.1  & 0.15 \\ \hline
			mIoU   & 52.8 & 55.7 & 55.3 & 53.6 \\ \hline
		\end{tabular}
		\caption{The influence of temperature parameter $\tau$.}
		\label{tab3}\vspace{-1.0em}
	\end{table}
	
	\begin{figure*}
		\hspace{-1.5em} \vspace{-2.0em}
		\centering 
		\includegraphics[scale=0.155]{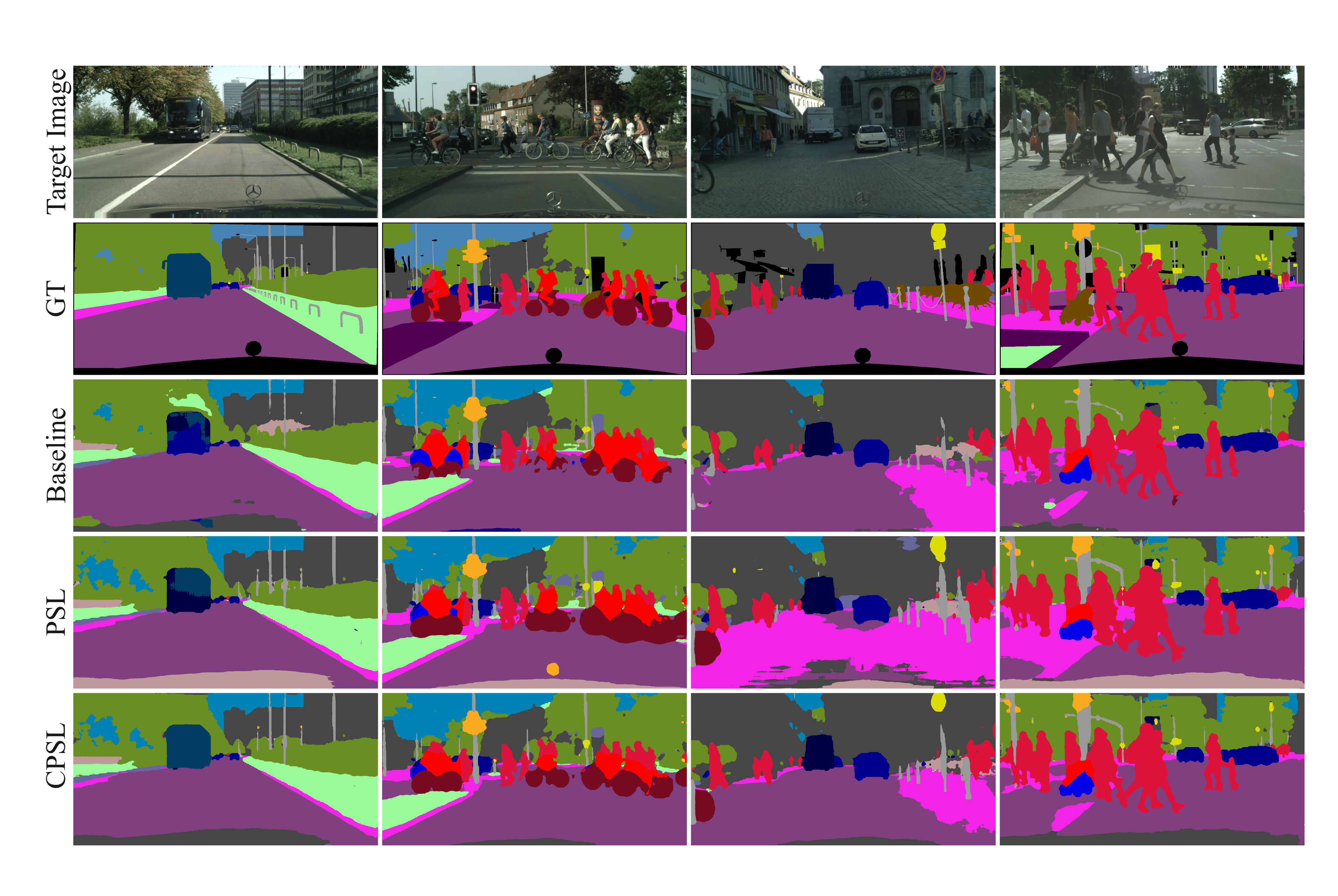}\\
		\caption{Qualitative results of PSL and CPSL on the GTA5$\to$Cityscapes task.}
		\label{fig1}\vspace{-2.0em}
	\end{figure*}
	
	\begin{figure*}[!t]
		\centering 
		\includegraphics[scale=0.24]{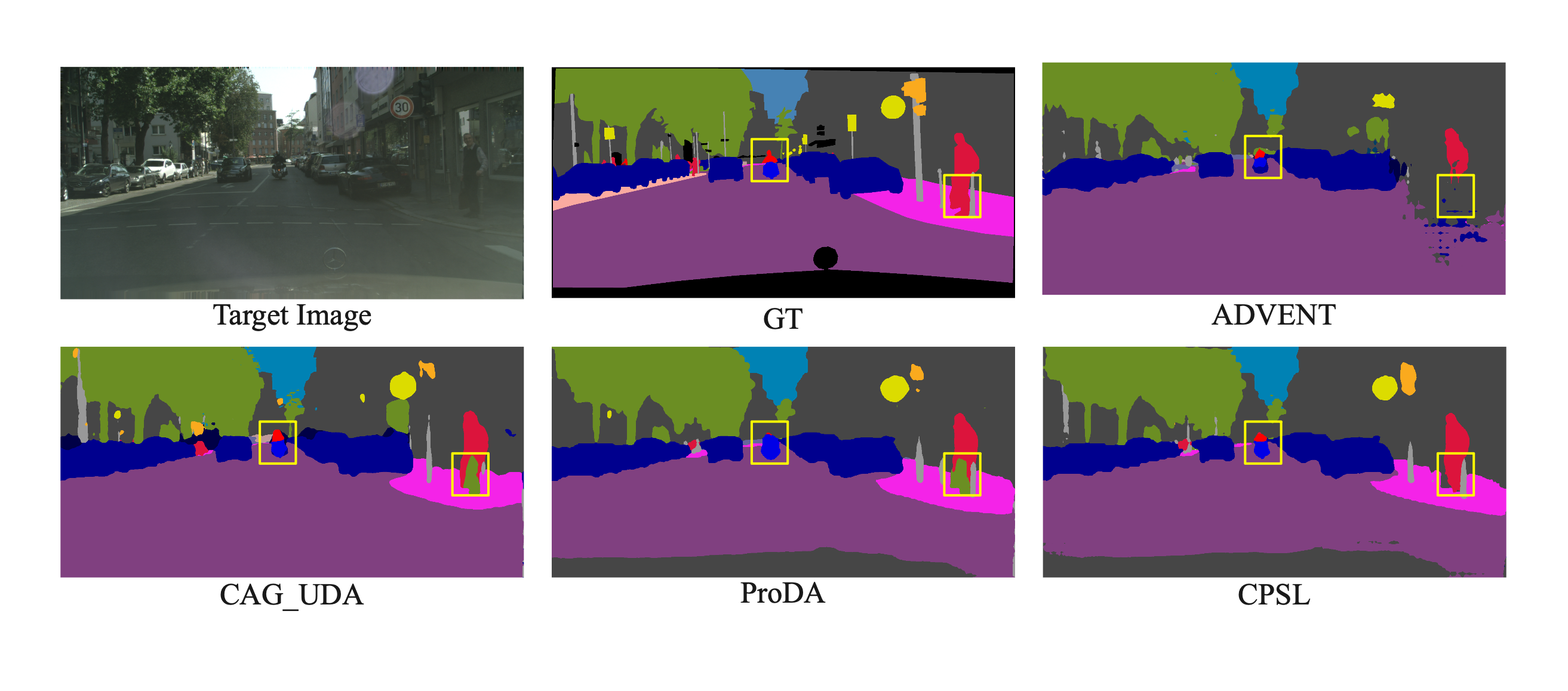}\\ \vspace{-2.5em}
		\caption{Qualitative comparison of different methods on the GTA5$\to$Cityscapes task.}
		\label{q1}\vspace{-1.5em}
	\end{figure*}
	\begin{figure*}[!t]
	\centering 
	\includegraphics[scale=0.24]{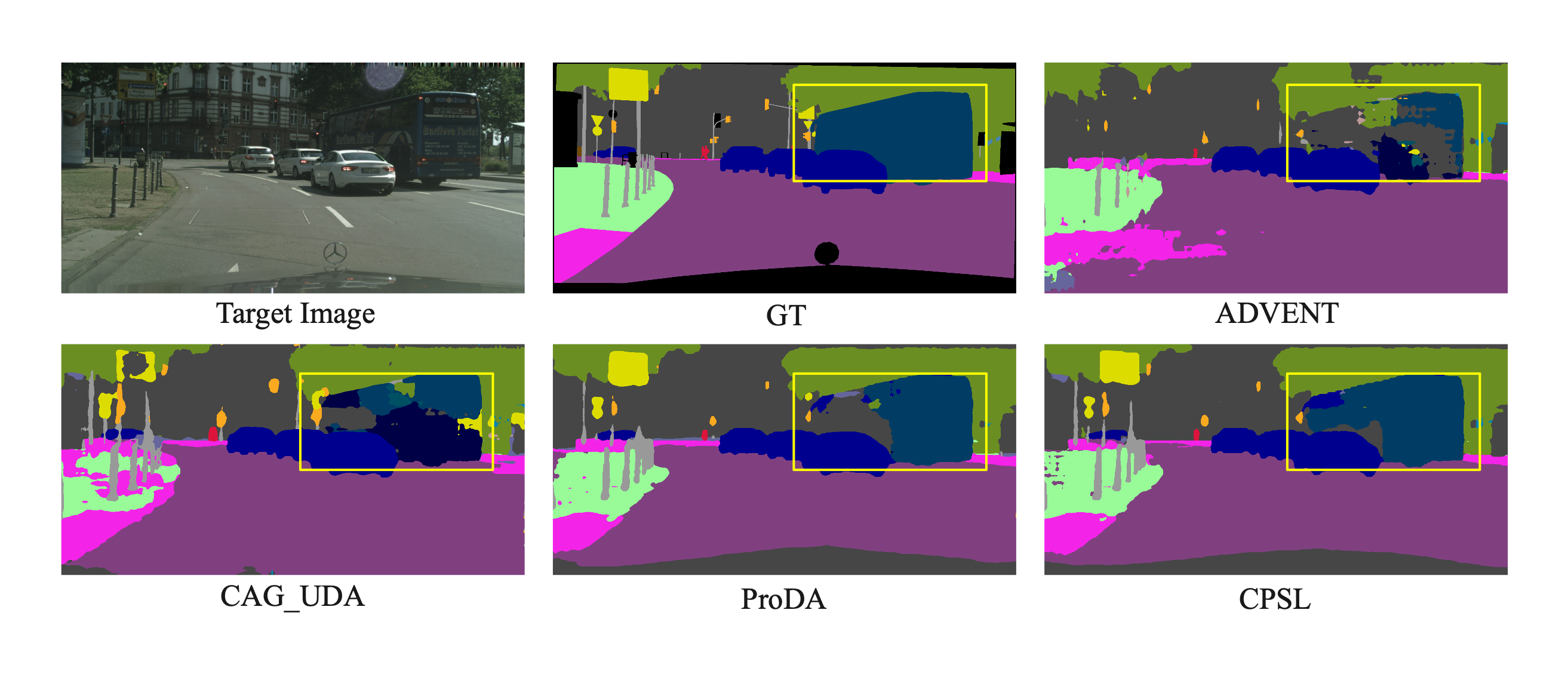}\\
	\vspace{-2.5em}
	\caption{Qualitative comparison of different methods on the GTA5$\to$Cityscapes task.}
	\label{q2}
\end{figure*}
	\subsection{Qualitative results}
	\textbf{PSL \textit{vs.} CPSL.} To better illustrate the performance of our method, we implement a variant of CPSL without class-balanced training, \ie, purely Pixel-level Self-Labeling (PSL). The qualitative results of PSL and CPSL are shown in Fig.~\ref{fig1}. Overall, CPSL is capable of producing more accurate segments across various scenes. Specifically, our method performs better on long-tailed categories, \eg “bus”, “bicycle”, “person”, “light”. Compared to PSL, the segment boundaries of CPSL tend to be clearer and closer to object boundaries, such as “bicycle” and “person”. Besides, it is noteworthy that PSL wrongly classifies the “road” class into the “sidewalk” class in a large area, which is attributed to the equipartition constraint applied on cluster assignments. This constraint is not useful and would even degrade the performance if the real class distribution is not uniform. However, this issue is solved by aligning class distribution of cluster assignments to that of pseudo labels.

\begin{figure*}[!t]
	\centering 
	\includegraphics[scale=0.24]{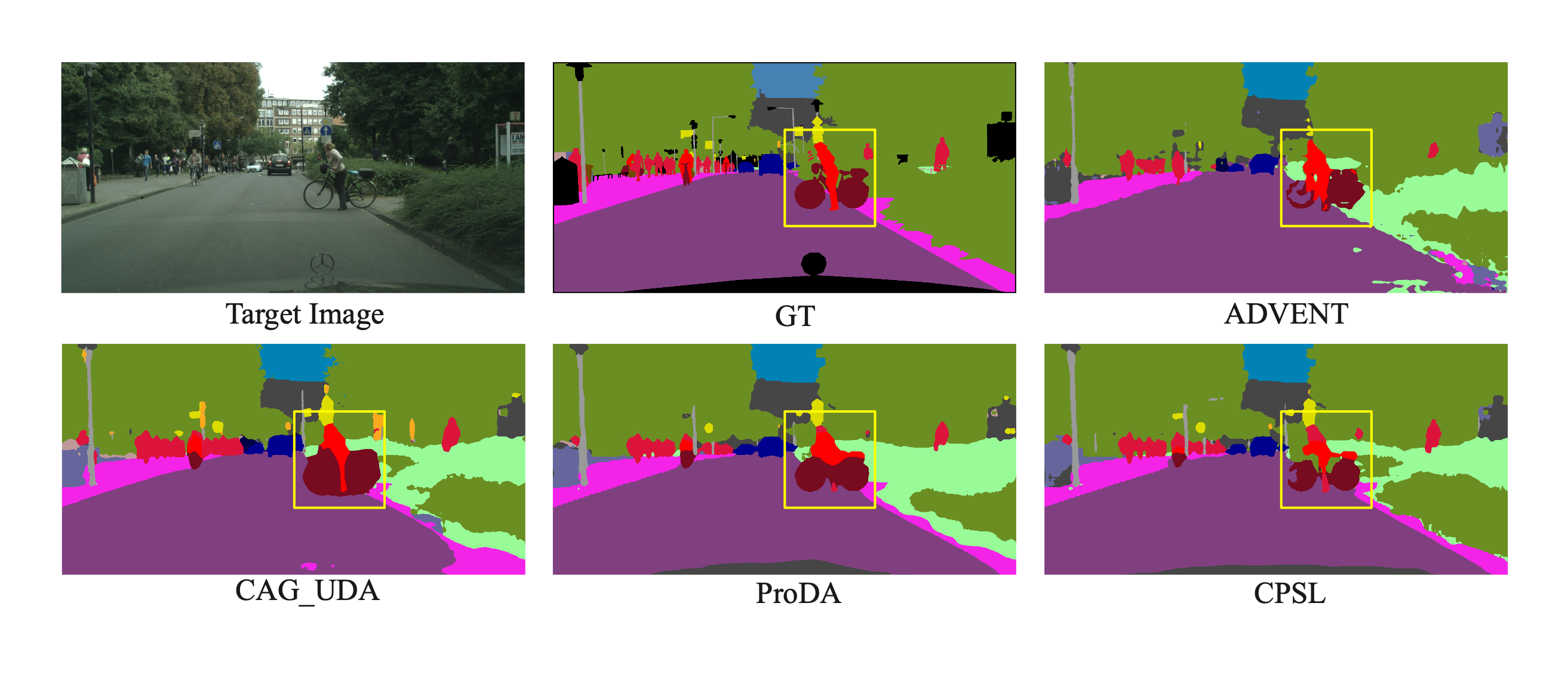}\\
	\vspace{-2.5em}
	\caption{Qualitative comparison of different methods on the GTA5$\to$Cityscapes task.}
	\label{q3}
\end{figure*}

\begin{figure*}[!t]
	\centering 
	\includegraphics[scale=0.24]{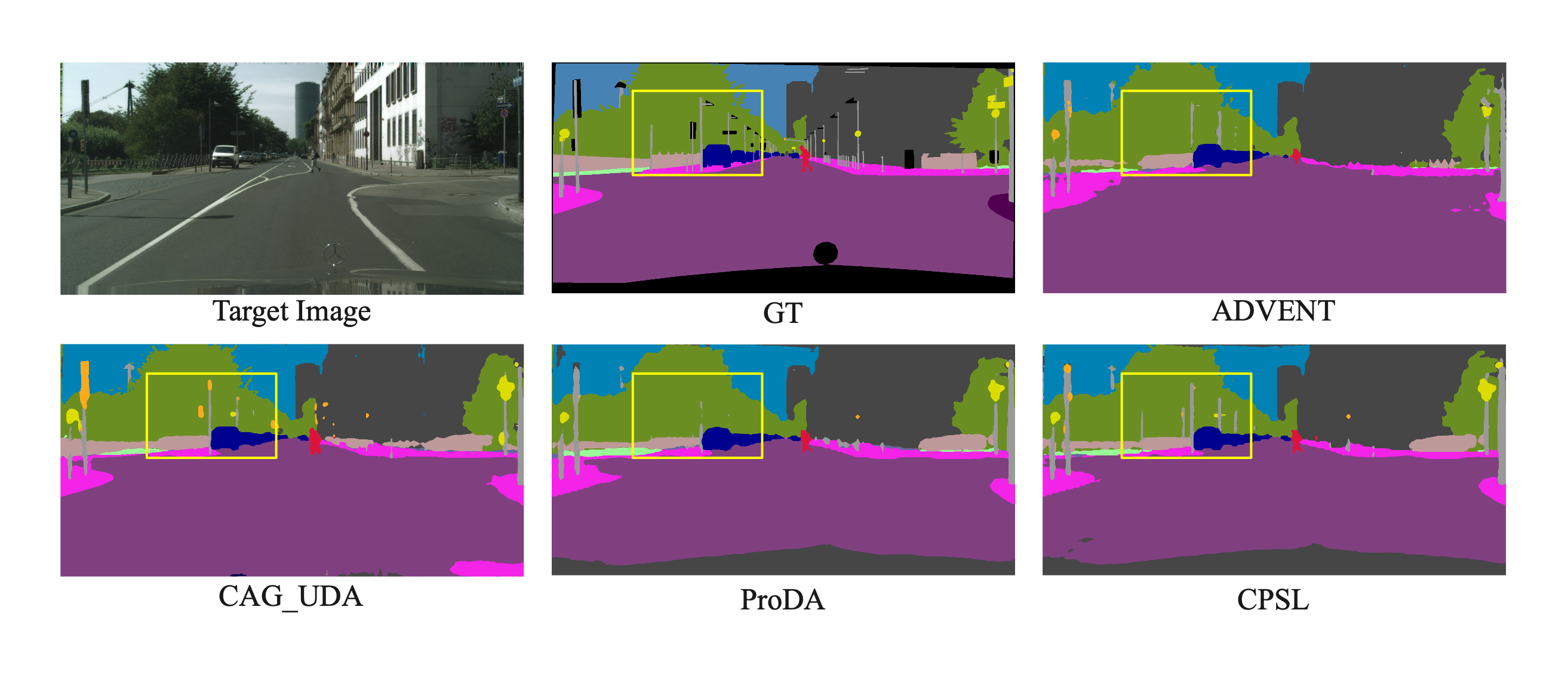}\\
	\vspace{-2.5em}
	\caption{Qualitative comparison of different methods on the GTA5$\to$Cityscapes task.}
	\label{q4}
\end{figure*}

\begin{figure*}[!t]
	\centering 
	\includegraphics[scale=0.24]{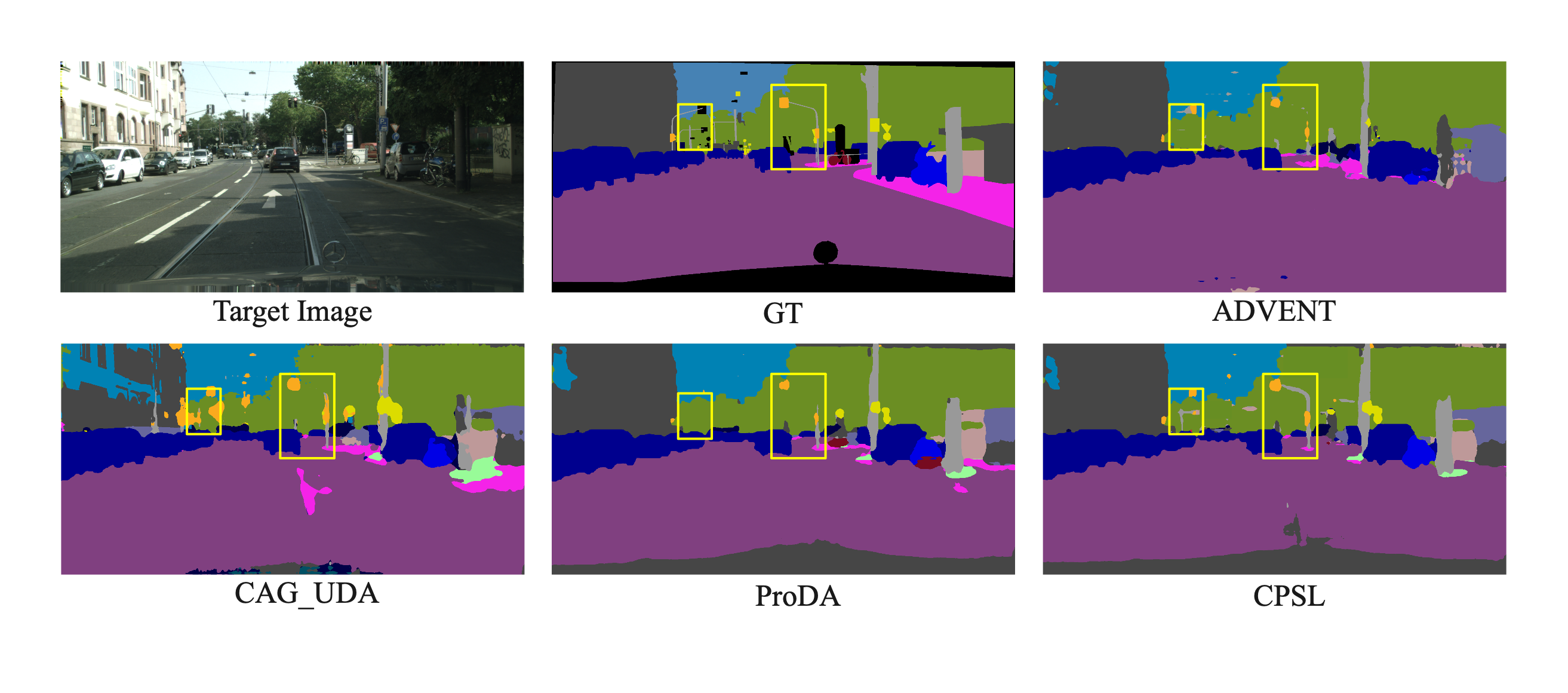}\\
	\vspace{-2.5em}
	\caption{Qualitative comparison of different methods on the GTA5$\to$Cityscapes task.}
	\label{q5}
\end{figure*}

\begin{figure*}[!t]
	\centering 
	\includegraphics[scale=0.24]{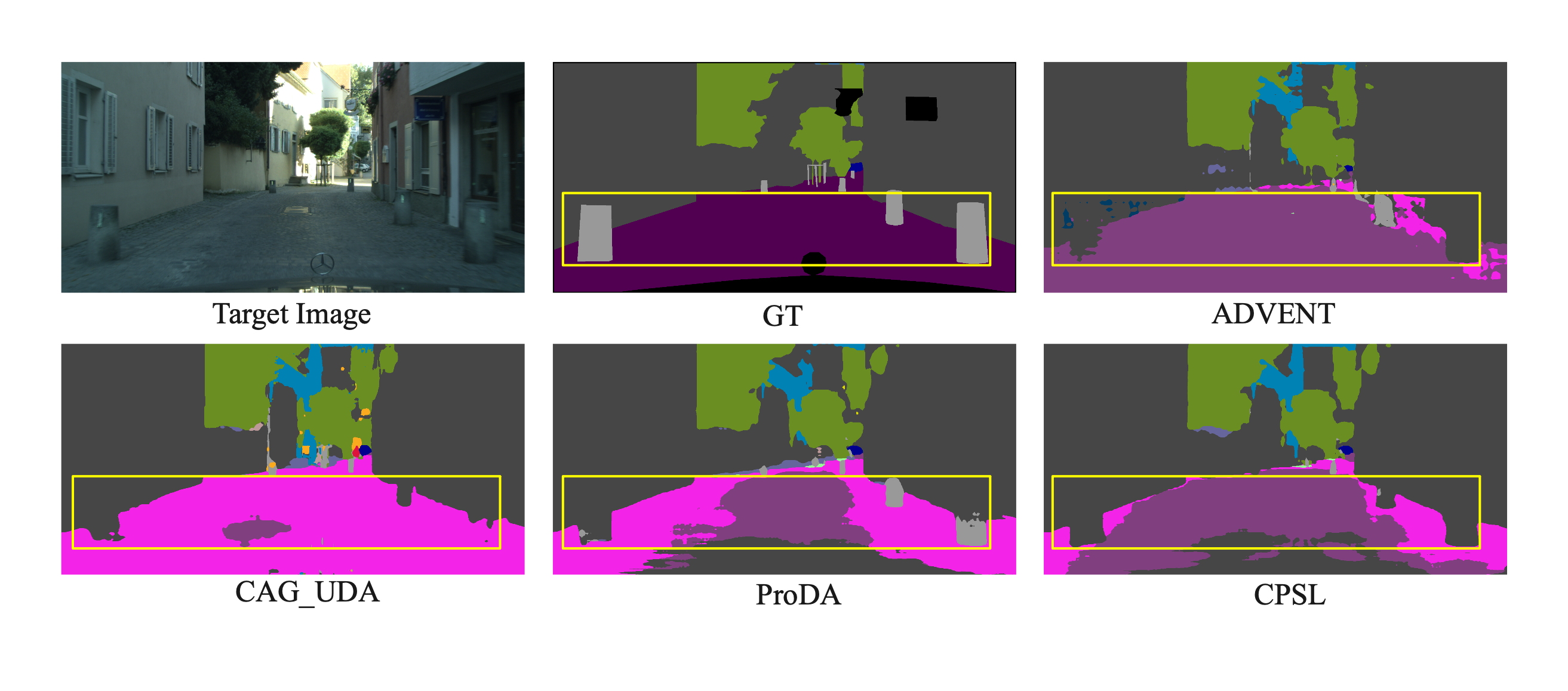}\\
	\vspace{-2.5em}
	\caption{Qualitative comparison of different methods on the GTA5$\to$Cityscapes task.}
	\label{q6}
\end{figure*}

\begin{figure*}[!t]
	\centering 
	\includegraphics[scale=0.24]{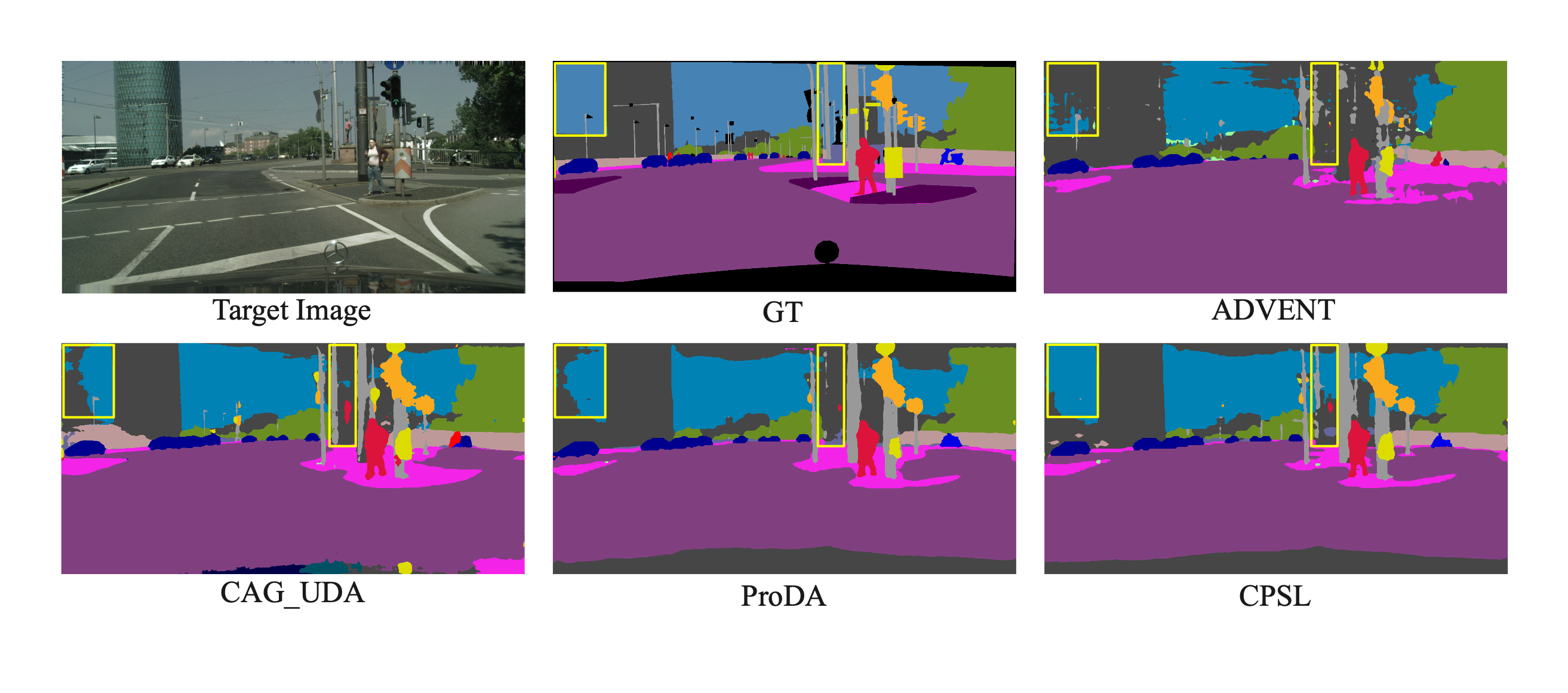}\\
	\vspace{-2.5em}
	\caption{Qualitative comparison of different methods on the GTA5$\to$Cityscapes task.}
	\label{q7}
\end{figure*}

	\textbf{Comparisons with state-of-the-arts.}
	As in Fig.~3 of the main manuscript, we compare our CPSL with other state-of-the-art methods. Here we provide more visualization results in Fig.~\ref{q1} - Fig.~\ref{q7}. Our method performs better on long-tailed categories, such as ``person", ``pole", ``traffic light", ``bus", and ``rider".

\end{document}